\documentclass[namedate,unnumsec,webpdf,contemporary,large,nocrop]{./oup-authoring-template}%

\graphicspath{%
  {./Figure/}%
}

\theoremstyle{thmstyleone}%
\theoremstyle{thmstyletwo}%
\theoremstyle{thmstylethree}%

\usepackage{booktabs}
\usepackage{multirow}
\usepackage{caption}
\usepackage{placeins}
\usepackage{cuted}
\usepackage{float}
\usepackage{tabularx}
\PassOptionsToPackage{hyphens}{url}
\usepackage{url}
\usepackage{hyperref}
\usepackage{algorithm}
\usepackage{algpseudocode}
\definecolor{revisionred}{RGB}{0,0,0}
\makeatletter
\DeclareRobustCommand{\rev}[1]{\begingroup\def\@linkcolor{revisionred}\def\@citecolor{revisionred}\def\@urlcolor{revisionred}\color{revisionred}#1\endgroup}
\makeatother
\newcommand{\revheading}[1]{\texorpdfstring{\textcolor{revisionred}{#1}}{#1}}
\hypersetup{
	colorlinks=true,
	linkcolor=black,
	urlcolor=blue,
	pdfborder={0 0 0}
}
\newcolumntype{Z}{>{\centering\arraybackslash}X}
\newcolumntype{Y}{>{\raggedright\arraybackslash}X}
\newcolumntype{C}[1]{>{\centering\arraybackslash}p{#1}}
\newcommand{\methodhead}[1]{\par\medskip\noindent\textbf{#1.}\quad}
\makeatletter
\renewcommand\subsubsection{\@startsection{subsubsection}{3}{\z@}{10pt plus 2pt minus 2pt}{4pt}{\normalfont\sffamily\small\bfseries}}
\makeatother

\makeatletter
\renewcommand*\CROP@@info{}
\def\shipout@PageObjects{}
\let\rest@dvi@pages\shipout@PageObjects
\makeatother
\AtBeginDocument{\crop[off]\hfuzz=\maxdimen\hbadness=10000}

\makeatletter
\def\boxtypetwohdstyle{\reset@font
\if@contemporary\sffamily\if@large\fontsize{9bp}{11.5bp}\selectfont\else\fontsize{9bp}{11bp}\selectfont\fi\bfseries\else
\if@traditional\fontsize{9bp}{11.75bp}\selectfont\else\fontsize{9bp}{11bp}\selectfont\fi\fi}
\makeatother

\begin{document}
\raggedbottom

\journaltitle{Bioinformatics}
\DOI{DOI added during production}
\copyrightyear{2026}
\pubyear{2026}
\vol{XX}
\issue{x}
\access{Published: Date added during production}
\appnotes{Paper}

\firstpage{1}

\title[CMGL: Confidence-guided Multi-omics Graph Learning]{CMGL: Confidence-guided Multi-omics Graph Learning for Cancer Subtype Classification}

\author[1]{Boyang Fan\ORCID{0009-0008-7291-7359}}
\author[2,$\ast$]{Hengchuang Yin\ORCID{0000-0002-9561-7817}}
\author[3]{Siyu Yi\ORCID{0000-0001-5124-2382}}
\author[4]{Yifan Wang\ORCID{0000-0001-7764-8698}}
\author[1]{Zhicheng Li\ORCID{0009-0003-0712-4225}}
\author[1]{Leijiyu Zhou\ORCID{0009-0006-3703-7378}}
\author[1]{Jiancheng Lv\ORCID{0000-0001-6551-3884}}
\author[5,$\ast$]{Wei Ju\ORCID{0000-0001-9657-951X}}

\address[1]{
	\orgdiv{College of Computer Science},
	\orgname{Sichuan University},
	\orgaddress{Chengdu, \postcode{610065}, \country{China}}
}
\address[2]{
	\orgdiv{Xinjiang Technical Institute of Physics and Chemistry},
	\orgname{Chinese Academy of Sciences},
	\orgaddress{Urumqi, \postcode{830011}, \country{China}}
}
\address[3]{
	\orgdiv{College of Mathematics},
	\orgname{Sichuan University},
	\orgaddress{Chengdu, \postcode{610065}, \country{China}}
}
\address[4]{
	\orgdiv{School of Artificial Intelligence and Data Science},
	\orgname{University of International Business and Economics},
	\orgaddress{Beijing, \postcode{100029}, \country{China}}
}
\address[5]{
	\orgdiv{School of Artificial Intelligence},
	\orgname{Sichuan University},
	\orgaddress{Chengdu, \postcode{610065}, \country{China}}
}

\corresp[$\ast$]{Corresponding authors: \href{mailto:juwei@scu.edu.cn}{juwei@scu.edu.cn}; \href{mailto:yinhengchuang@ms.xjb.ac.cn}{yinhengchuang@ms.xjb.ac.cn}.}

\received{Date}{0}{Year}
\revised{Date}{0}{Year}
\accepted{Date}{0}{Year}

\abstract{
\textbf{Motivation:} Multi-omics integration can improve cancer subtyping, but modality informativeness and noise vary across cancer types and patients. \rev{Most graph methods for multi-omics data learn modality contributions within the downstream classification objective, leaving predictive reliability for each patient implicit. As a result, uninformative modalities can weaken the fused representation, while unreliable omics can introduce noisy patient relationships into graph propagation. To address these two problems, we propose CMGL, which produces a separate reliability estimate before fusion and uses consensus patient neighborhoods for graph classification.}\\[2pt]
\textbf{Results:} \rev{CMGL estimates modality confidence for each patient through evidential deep learning, fixes these values during fusion across omics, and performs classification on an independently specified consistency graph.} On four MLOmics cancer-subtype tasks and the 32-class pan-cancer task, CMGL consistently improves over the strongest baseline, surpassing it by 4.03\% in average accuracy on the four single-cancer tasks. Its representations recover the PAM50 intrinsic subtypes of breast invasive carcinoma (BRCA), and the model trained on BRCA transfers without fine tuning to kidney renal clear cell carcinoma (KIRC), stratifying patients into prognostically distinct groups.\\
\textbf{Availability:} \rev{The code is available at \url{https://doi.org/10.5281/zenodo.21921069}.}\\
\textbf{Contact:} \href{mailto:juwei@scu.edu.cn}{juwei@scu.edu.cn} (Wei Ju); \href{mailto:yinhengchuang@ms.xjb.ac.cn}{yinhengchuang@ms.xjb.ac.cn} (Hengchuang Yin)\\
}
\keywords[\keywordsname]{multi-omics integration, cancer subtype classification, evidential deep learning, graph neural networks, uncertainty-aware learning}

\maketitle

\section{Introduction}

Molecular classification of cancer subtypes based on multi-omics profiling has become central to precision oncology \citep{Ritchie2015,Hasin2017}. By integrating mRNA expression, miRNA expression, DNA methylation, and copy number variation, these approaches guide patient stratification and treatment selection. In breast invasive carcinoma (BRCA), for instance, the PAM50 intrinsic subtypes defined through transcriptomic profiling have directly informed clinical decision-making for over a decade \citep{TCGA2012brca,Perou2000}. However, the informativeness of each omics modality varies across cancer types: BRCA subtypes are defined primarily by mRNA expression and DNA methylation, whereas ovarian serous cystadenocarcinoma (OV) subtypes are driven primarily by copy-number alterations \citep{TCGA2011ov}. \rev{Comparative evaluations show that the contribution of an omics modality depends on the cancer and dataset, and that adding an uninformative view does not necessarily improve cancer subtyping \citep{Duan2021multiomics}. Studies of integration under noisy conditions further identify modality-specific noise and data-specific patterns as obstacles to learning a consistent multi-omics representation \citep{Wang2021defusion}.} \rev{This variation across cancers motivates an assessment of omics reliability for each patient before the modality representations are combined for subtype prediction.}

Multi-omics fusion has progressed from feature-selection and kernel approaches \citep{Rappoport2018,Zhao2017} to deep methods such as autoencoders and attention mechanisms \citep{Picard2021,Li2018,Cheerla2019}, capsule networks \citep{Sabour2017,Zhang2025}, and pretrained multi-omics transformers \citep{Wang2024tmonet,SubtypeMGTP2024}. \rev{Complementary lines of work include similarity network fusion \citep{Wang2014snf}, integration for incomplete or denoised data \citep{Rappoport2019nemo,Wang2021defusion}, and general multimodal representation learning \citep{Baltrusaitis2019multimodal}.} More recently, graph-based patient-relationship models \citep{Wang2021mogonet,Wu2024mosgat,DGHNN2025,HallmarkGraph2025} have become a leading direction. Among these, MOGONET \citep{Wang2021mogonet} constructs an independent GCN for each omics modality and fuses the per-view predictions through a view-correlation discovery network (VCDN), while MOSGAT \citep{Wu2024mosgat} introduces a specificity-aware graph-attention network combined with cross-modal attention. For uncertainty modeling, evidential deep learning (EDL) \citep{Sensoy2018,Josang2016}, prior networks \citep{Malinin2018}, and trusted multi-view classification \citep{Han2023} provide tools for obtaining predictions and uncertainty estimates in a single forward pass, and HTML \citep{Lu2023html} applies Dirichlet evidence fusion to multi-omics cancer classification. Nevertheless, these methods generally treat modality weights as latent variables jointly optimized with the classification objective, so the resulting weights reflect classification-driven optima rather than modality-level reliability estimates. \rev{This leaves predictive reliability for each patient unclear when fusion and graph classification are optimized within the same downstream objective.}

\begin{figure*}[!t]
	\centering
	\includegraphics[width=\textwidth,trim=0 80 0 0,clip]{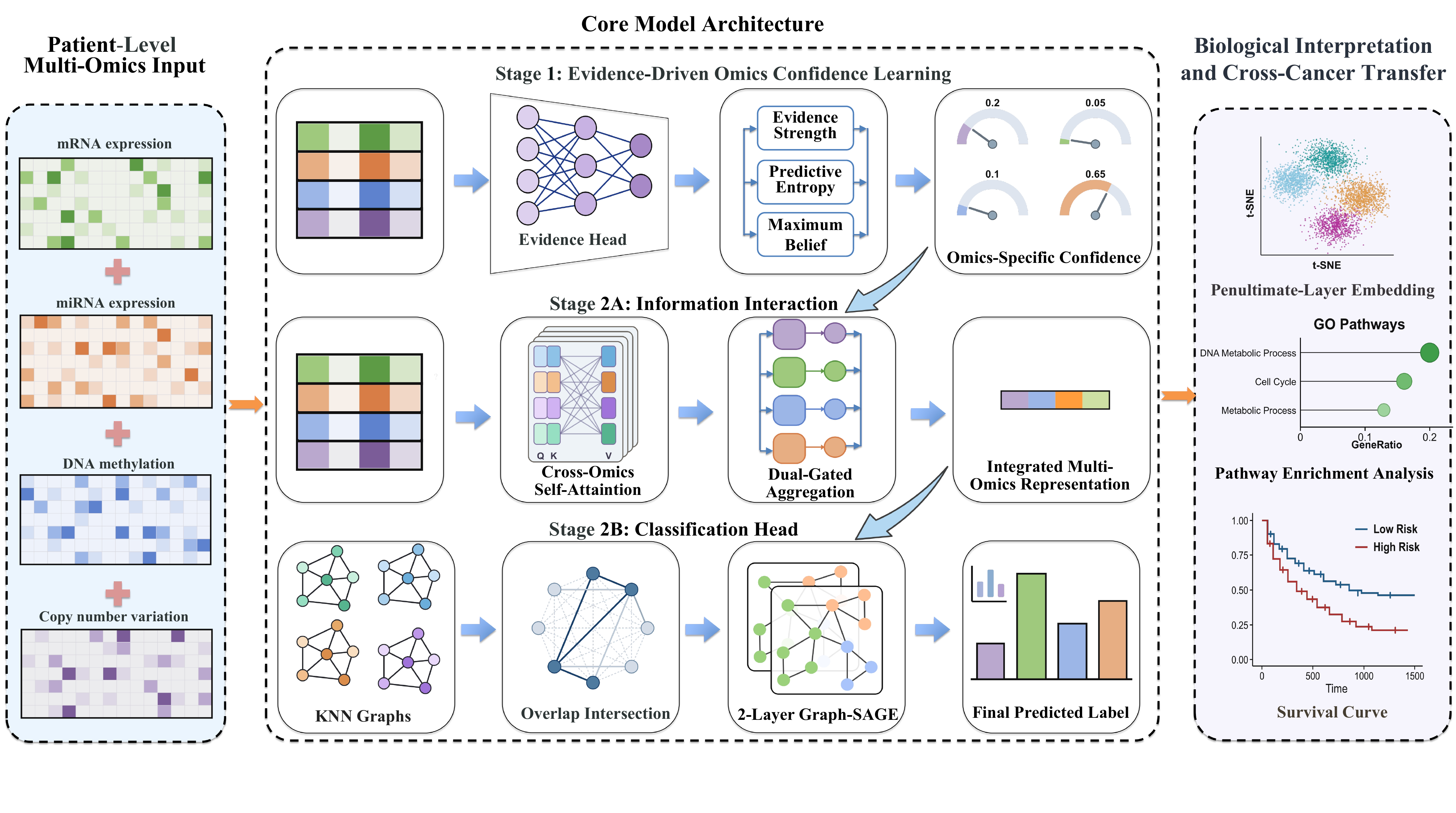}
\caption{Overview of CMGL. Stage~1 takes four omics modalities of each patient and passes them through per-modality encoders and evidential heads to obtain per-sample modality confidence scores, which are then frozen. \rev{Stage~2 uses the fixed confidence values to gate cross-omics fusion, while an independently specified consistency intersection $k$-NN graph defines patient neighborhoods.} A two-layer GraphSAGE with a residual link from $\mathbf{z}_i$ to the output then propagates subtype information on this graph to produce the final classification.}
	\label{fig:cmgl_overview}
\figalttext[CMGL architecture and analysis workflow]{Alt text: Flow diagram of CMGL from four omics inputs through modality confidence learning for each patient, confidence-gated fusion, intersection-graph classification, and downstream biological analyses.}%
\end{figure*}

\rev{Modality-specific noise can degrade the shared representation learned across omics \citep{Wang2021defusion}, while structural noise in a patient graph can impair GNN message passing \citep{Dong2023graphnoise}.} Biologically, this means that the molecular signatures of true cancer subtypes can be diluted by noise from uninformative modalities. The core question is therefore how to identify which omics modalities are trustworthy for each patient before fusion and graph reasoning, so that reliable modalities shape subsequent information flow. \rev{Inspired by assessments of view trustworthiness for individual samples \citep{Han2023,Zhang2026prest}, CMGL estimates predictive confidence before fusion and uses patient relationships supported across modality graphs, with these operations regulating feature aggregation and information propagation between patients separately.}

In this work, we propose CMGL (Confidence-guided Multi-omics Graph Learning), which is motivated by the observation that modality reliability should be estimated independently before fusion and graph reasoning. \rev{To our knowledge, among the graph-based multi-omics cancer subtyping methods discussed above, CMGL is the first to learn modality confidence for each patient in a separate stage and keep it fixed during downstream fusion.} CMGL adopts a two-stage design. \rev{Whereas MOGONET and MOSGAT learn view fusion within the classifier and HTML combines modality-level Dirichlet evidence, CMGL uses the fixed confidence signal for feature-level fusion while specifying patient neighborhoods independently.}

We evaluate CMGL on four cancer subtype classification tasks from the MLOmics benchmark \citep{Yang2025mlomics} derived from The Cancer Genome Atlas (TCGA). \rev{The evaluation further includes a pan-cancer benchmark with 32 classes and BRCA to KIRC transfer, covering predictive performance and biological structure.}

\section{Materials and Methods}

\subsection{Data Sources}

We build our experiments upon the MLOmics multi-omics cancer database \citep{Yang2025mlomics}, which curates and standardizes data from TCGA \citep{Hoadley2018} to provide a benchmark for machine learning research on multi-omics classification. The primary benchmark comprises four single-cancer subtype classification tasks: BRCA \citep{TCGA2012brca}, glioblastoma multiforme (GBM) \citep{TCGA2008gbm}, lower-grade glioma (LGG) \citep{TCGA2015lgg}, and OV \citep{TCGA2011ov}. The pan-cancer task is an independent benchmark. In addition, we introduce KIRC as an unlabeled cross-cancer transfer cohort for forward-only (frozen) inference and downstream prognostic stratification. For each patient we use the same four omics modalities: mRNA expression, miRNA expression, DNA methylation, and copy number variation (CNV). For the supervised benchmark tasks, subtype labels are taken directly from curated GDC/TCGA annotations packaged in MLOmics, whereas KIRC is used without target-domain subtype labels and is associated only with clinical survival endpoints for transfer analysis. Dataset statistics, including sample sizes, subtype classes, and per-modality feature dimensionalities, are provided in Table~\ref{tab:dataset_overview}.

\begin{table*}[!htbp]
	\centering
	\caption{Dataset overview. Sample counts, subtype classes, and per-modality feature dimensionality for the five MLOmics benchmark tasks used in this study and the KIRC transfer cohort.}
	\label{tab:dataset_overview}
	\small
	\begin{tabular*}{\textwidth}{@{\extracolsep\fill}lcccccc@{\extracolsep\fill}}
		\specialrule{1.0pt}{0pt}{0pt}
		Task & Samples & Classes & mRNA dim & miRNA dim & Methylation dim & CNV dim \\
		\specialrule{0.6pt}{0pt}{0pt}
		BRCA & 671 & 5 & 5000 & 200 & 5000 & 5000 \\
		GBM & 244 & 5 & 5000 & 200 & 5000 & 5000 \\
		LGG & 247 & 3 & 5000 & 200 & 5000 & 5000 \\
		OV & 284 & 4 & 5000 & 200 & 5000 & 5000 \\
		Pan-cancer & 8314 & 32 & 3217 & 383 & 3139 & 3105 \\
		KIRC (transfer cohort) & 314 & N/A\,$^{\dagger}$ & 5000 & 200 & 5000 & 5000 \\
		\specialrule{0.6pt}{0pt}{0pt}
		\multicolumn{7}{l}{\scriptsize $^{\dagger}$~KIRC is used without target-domain subtype labels.} \\
		\specialrule{1.0pt}{0pt}{0pt}
	\end{tabular*}
\end{table*}

\subsection{Overall Framework}

\rev{To estimate reliability for each patient before fusion and graph reasoning, we designed CMGL with two stages (Fig.~\ref{fig:cmgl_overview}).} Stage~1 (Section~2.3) independently estimates, for each patient and each omics modality, how reliable that modality is for classification. Stage~2 (Section~2.4) performs cross-omics feature fusion, consistency graph construction, and graph-based classification under the frozen confidence scores from Stage~1. The key design principle is that reliability estimation precedes fusion and graph reasoning, and is not updated by downstream gradients.

\rev{Let the dataset be $\mathcal{D}=\{(\mathbf{x}_i^{(1)},\ldots,\mathbf{x}_i^{(M)},y_i)\}_{i=1}^{N}$, where $y_i\in\{1,\ldots,C\}$ is the class label and $M$ is the number of modalities.} For each patient, CMGL produces modality confidence $\mathbf{r}_i$, a fused representation $\mathbf{z}_i$, and classification probability $\hat{\mathbf{p}}_i$. \rev{Complete notation, including modality confidence $r_i^{(m)}$, fused representation $\mathbf z_i$, intersection graph $\mathcal E_{\cap}$, and final prediction $\hat{\mathbf p}_i$, can be found in Supplementary Table~\ref{tab:method_notation}.}

\subsection{Stage 1: Evidence-driven Omics Confidence Learning}

The goal of Stage~1 is to answer, for each patient and each omics modality, how much that modality should be trusted for classification. This is necessary because modality informativeness varies both across cancer types (Section~1) and across patients within a single cancer type, due to batch effects and tumor heterogeneity. \rev{We estimate confidence before joint representation learning, and Stage~2 uses it to allocate each patient's modality contributions during fusion.}

\methodhead{Evidence Modeling}
For the $m$-th omics modality, an encoder $E_m^{(r)}(\cdot)$ maps the raw input into a unified hidden space, and an evidence head outputs non-negative class-wise evidence $\mathbf{e}_i^{(m)}\in\mathbb{R}_{\ge 0}^{C}$. Following the EDL framework \citep{Sensoy2018}, we construct Dirichlet parameters and the total Dirichlet strength
\begingroup
\color{revisionred}
\begin{equation}
\boldsymbol{\alpha}_i^{(m)} = \mathbf{e}_i^{(m)} + \mathbf{1}_C,
\label{eq:dirichlet}
\end{equation}
\endgroup
\begin{equation}
S_i^{(m)} = \sum_{c=1}^{C} \alpha_{i,c}^{(m)},
\label{eq:dirichlet_strength}
\end{equation}
which yield the Dirichlet predictive mean
\begingroup
\color{revisionred}
\begin{equation}
\boldsymbol{\pi}_i^{(m)} = \boldsymbol{\alpha}_i^{(m)} / S_i^{(m)},
\label{eq:belief}
\end{equation}
\endgroup
and the epistemic uncertainty
\begin{equation}
u_i^{(m)} = C / S_i^{(m)}.
\label{eq:uncertainty}
\end{equation}
Unlike Monte Carlo Dropout \citep{Gal2016} and deep ensembles \citep{Lakshminarayanan2017}, which require multiple forward passes, EDL obtains predictions and uncertainty estimates in a single pass; post-hoc calibration approaches \citep{Guo2017,Naeini2015} also require a separate recalibration pass.

\methodhead{Quality Estimator}
A single uncertainty signal cannot robustly characterize modality reliability. \rev{CMGL therefore extracts four complementary quality signals. The log evidence strength $a_i^{(m)}=\log(1+\sum_c e_{i,c}^{(m)})$ measures accumulated evidence. The epistemic uncertainty $u_i^{(m)}$ reflects uncertainty in the model's knowledge. The normalized predictive entropy $\widetilde H_i^{(m)}=-\sum_c\pi_{i,c}^{(m)}\log\pi_{i,c}^{(m)}/\log C$ measures ambiguity in the class distribution. The maximum predictive probability $\pi_{i,\max}^{(m)}=\max_c \pi_{i,c}^{(m)}$ indicates concentration on the leading class. Together, these signals characterize modality specific predictions from complementary aspects.} \rev{A scoring network $Q_m(\cdot)$ for each modality maps these quantities to $M$ scores. A learned positive temperature factor $T$ then scales the scores, and softmax normalization yields the modality confidence:}
\begingroup
\color{revisionred}
\begin{equation}
r_i^{(m)} = \frac{\exp\!\left( Q_m\!\left(a_i^{(m)},u_i^{(m)},\widetilde H_i^{(m)},\pi_{i,\max}^{(m)}\right) / T \right)}{\sum_{m'=1}^{M} \exp\!\left( Q_{m'}\!\left(a_i^{(m')},u_i^{(m')},\widetilde H_i^{(m')},\pi_{i,\max}^{(m')}\right) / T \right)}.
\label{eq:confidence}
\end{equation}
\endgroup
Intuitively, a modality receives high confidence when it produces concentrated evidence for a single class and low entropy across classes.

\methodhead{Stage 1 Loss}
\rev{The Stage~1 objective combines three complementary terms: an EDL loss $\mathcal{L}_{\mathrm{edl}}$ with an annealed KL regularizer for evidence learning, a confidence-weighted classification loss $\mathcal{L}_{\mathrm{cls}}^{(r)}$ that reduces the contribution of low confidence modality predictions to label discrimination, and a diversity constraint $\mathcal{L}_{\mathrm{div}}$ that discourages uniform weighting across modalities. Their weights are $\lambda_{\mathrm{edl}}$, $\lambda_{\mathrm{cls}}^{(r)}$, and $\lambda_{\mathrm{div}}$, respectively:}
\begin{equation}
\mathcal{L}_{\mathrm{stage1}} = \lambda_{\mathrm{edl}}\,\mathcal{L}_{\mathrm{edl}} + \lambda_{\mathrm{cls}}^{(r)}\,\mathcal{L}_{\mathrm{cls}}^{(r)} + \lambda_{\mathrm{div}}\,\mathcal{L}_{\mathrm{div}}.
\label{eq:stage1_loss}
\end{equation}
After training, the confidence vector $\mathbf r_i$ is frozen and used by Stage~2.

\subsection{Stage 2: Confidence-guided Cross-omics Feature Fusion}

\rev{With the confidence values fixed in Stage~1, Stage~2 combines complementary molecular signals while preserving the reliability assigned to each modality for each patient.}

\methodhead{Cross-omics Attention}
\rev{To place modalities in a shared space while keeping their sources distinct, independent encoders $E_m^{(f)}(\cdot)$ map each modality and add a learnable modality identity: $\mathbf v_i^{(m)}=E_m^{(f)}(\mathbf x_i^{(m)})+\mathbf o_m$ \citep{Vaswani2017}.} \rev{Multi-head self-attention then exchanges information across the $M$ modality tokens and maps each token to $\widetilde{\mathbf v}_i^{(m)}$, allowing the attention weights to identify complementary features across modalities.}

\methodhead{Confidence-guided Gated Aggregation}
\rev{Multiplicative gating at the sample level applies the Stage~1 reliability signal during fusion, and it operates at both the feature and modality levels \citep{Arevalo2017gmu}.} A per-dimension gating vector
\begingroup
\color{revisionred}
\begin{equation}
\mathbf{g}_i^{(m)} = \sigma\!\left( \mathbf{W}_g\, \bigl[\widetilde{\mathbf{v}}_i^{(m)} \,\|\, r_i^{(m)}\bigr] + \mathbf b_g \right),
\label{eq:gate}
\end{equation}
\endgroup
modulates individual feature dimensions, while the scalar confidence $r_i^{(m)}$ allocates omics-level weights. The fused representation is
\begingroup
\color{revisionred}
\begin{equation}
\mathbf{z}_i = \sum_{m=1}^{M} r_i^{(m)} \bigl(\mathbf{g}_i^{(m)} \odot \widetilde{\mathbf{v}}_i^{(m)}\bigr).
\label{eq:fused}
\end{equation}
\endgroup
Low-confidence modalities are suppressed at both levels simultaneously, preventing noisy omics from diluting subtype-discriminative features.

\methodhead{\rev{Consistency Intersection Graph Construction}}
\rev{After Stage~2 forms a representation weighted by reliability, CMGL separately limits message passing to neighborhoods supported across molecular views. This design is motivated by evidence that graph changes and structural noise can distort GNN aggregation \citep{JU2024106207,Dong2023graphnoise}.} \rev{CMGL constructs a $k$-NN graph $\mathcal E^{(m)}$ for each modality using cosine distance in the standardized input feature space and retains their intersection:}
\begin{equation}
\mathcal{E}_{\cap} = \bigcap_{m=1}^{M} \mathcal{E}^{(m)}.
\label{eq:intersection}
\end{equation}
\rev{Together, these steps apply the learned modality reliability during feature fusion, and use neighborhood consistency across views to govern graph propagation.}

\methodhead{Transductive Edge Policy}
Edges are only built between training nodes or between a training node and an evaluation (validation or test) node; edges between two evaluation nodes are never built, so each evaluation prediction aggregates only from training neighbors. \rev{The validation and test graphs are constructed independently from their respective covariates, so covariates from evaluation nodes participate in graph construction. Validation labels are used only for model selection, whereas test labels are reserved for final evaluation.}

\methodhead{Adaptive Neighbor Selection}
\rev{For each task and fold a short Stage~2 warm up independently selects $k^\ast$ from $\mathcal K=\{7,11,15,19,23\}$ on the fold's validation split using Macro-F1, and assigns ties to the larger $k$ for the corresponding test evaluation.}

\methodhead{\rev{Graph Message Passing and Classification}}
\rev{The consensus graph defines the neighborhood structure, and GraphSAGE combines each patient's fused representation with subtype information from supported neighbors.} \rev{For the stacked feature matrix $\mathbf Z$, the residual embedding $\mathbf h_i$ and class prediction $\hat{\mathbf p}_i$ are defined as}
\begingroup
\color{revisionred}
\begin{equation}
\begin{aligned}
\mathbf h_i &= \mathrm{GraphSAGE}_{2}\!\left(\mathbf Z,\mathcal E_{\cap}\right)_i + \mathbf W_{\mathrm{res}}\mathbf z_i,\\
\hat{\mathbf{p}}_i &= \mathrm{softmax}(\mathbf{W}_{\mathrm{cls}}\mathbf h_i + \mathbf{b}_{\mathrm{cls}}).
\end{aligned}
\label{eq:classifier}
\end{equation}
\endgroup
\rev{The embeddings $\mathbf h_i$ are used for all downstream biological analyses.}

\methodhead{Stage 2 Loss}
The Stage~2 objective combines two terms. A label-smoothing cross-entropy loss \citep{Muller2019} with $\epsilon=0.1$ and square-root inverse-frequency class weights normalized to have mean one provides the classification signal,
\begingroup
\color{revisionred}
\begin{equation}
\mathcal{L}_{\mathrm{ce}} = -\frac{1}{|\mathcal B|}\sum_{i\in\mathcal B}w_{y_i}\sum_{c=1}^{C}\widetilde{y}_{i,c}\log \hat{p}_{i,c},
\label{eq:ce}
\end{equation}
\endgroup
\begingroup
\color{revisionred}
\begin{equation}
\widetilde{y}_{i,c} = (1-\epsilon)\,\mathbb{I}[c=y_i] + \frac{\epsilon}{C-1}\mathbb{I}[c\neq y_i],
\label{eq:label_smoothing}
\end{equation}
\endgroup
\rev{We compute a supervised contrastive loss \citep{Khosla2020,Chen2020simclr} with temperature $\tau=0.1$ over the normalized embeddings $\widetilde{\mathbf h}_i=\mathbf h_i/\|\mathbf h_i\|_2$, bringing embeddings from the same class closer together and those from different classes farther apart. For $\mathcal P(i)=\{j\in\mathcal B\setminus\{i\}:y_j=y_i\}$ and $\mathcal B_+=\{i\in\mathcal B:|\mathcal P(i)|>0\}$, the loss averages over valid anchors in the training node set:}
\begingroup
\color{revisionred}
\begin{equation}
\mathcal{L}_{\mathrm{supcon}} = -\frac{1}{|\mathcal B_+|}\sum_{i\in\mathcal B_+}\frac{1}{|\mathcal P(i)|}\sum_{j\in\mathcal P(i)}\log\frac{\exp(\widetilde{\mathbf h}_i\cdot\widetilde{\mathbf h}_j/\tau)}{\sum_{\ell\in\mathcal B\setminus\{i\}}\exp(\widetilde{\mathbf h}_i\cdot\widetilde{\mathbf h}_{\ell}/\tau)},
\label{eq:supcon}
\end{equation}
\endgroup
The Stage~2 loss is
\begin{equation}
\mathcal{L}_{\mathrm{stage2}} = \lambda_{\mathrm{cls}}\,\mathcal{L}_{\mathrm{ce}} + \lambda_{\mathrm{con}}\,\mathcal{L}_{\mathrm{supcon}}.
\label{eq:stage2_loss}
\end{equation}

\rev{After Stage~2 optimization under the fixed confidence scores, validation Macro-F1 selects the checkpoint used for subsequent classification and biological interpretation tasks.}

\section{Results}

\subsection{Experimental Settings}

\rev{To evaluate the CMGL framework under consistent data partitions, benchmark experiments follow 5-fold stratified cross-validation. In each fold, 70\% of each cohort is used for training, 10\% for validation, and 20\% for testing.} For the small-sample generalization experiments, training sets are randomly subsampled within each fold at fractions 1.0, 0.5, and 0.3. The four primary tasks report Accuracy, Macro-F1, Weighted-F1, Macro-Recall, Macro-AUC, and Macro-AUPRC; the pan-cancer task reports the same metric set. \rev{Paired comparisons use three seeds and five folds. We apply Wilcoxon tests on signed ranks with two-sided alternatives, followed by the Benjamini-Hochberg correction. All $p$ values for these comparisons use this setting, and complete results are provided in Supplementary Table~\ref{tab:cover_all_baseline_fold_tests}.}

For biological interpretability we use frozen inference on the best-performing BRCA fold model selected by validation Macro-F1: a single forward pass with no parameter updates. The penultimate-layer embeddings and predictions from all samples are then used for marker gene discovery, pathway enrichment, survival analysis, and t-SNE visualization \citep{vanderMaaten2008}.

We compare against four baselines spanning three technical directions: MOGONET \citep{Wang2021mogonet} (multi-graph GCN with VCDN fusion), MOSGAT \citep{Wu2024mosgat} (specificity-aware GAT with cross-modal attention), HTML \citep{Lu2023html} (Dirichlet evidence with Dempster-Shafer fusion), and MOCapsNet \citep{Zhang2025} (capsule networks with self-attention fusion). All baselines use their respective default hyperparameters or official implementations.

CMGL employs an input hidden dimension of 128, a two-layer GraphSAGE with output hidden dimensions $128\!\to\!64$ (yielding 64-dimensional penultimate-layer embeddings), and a consistency intersection graph construction. Stages~1 and~2 are trained independently, with Stage~2 operating under frozen confidence priors. The default operating point for the main benchmark uses Stage-1 weights $\lambda_{\mathrm{edl}}{=}1.5$ with annealing step $50$, $\lambda_{\mathrm{cls}}^{(r)}{=}1.5$, $\lambda_{\mathrm{div}}{=}1.0$, and Stage-2 weights $\lambda_{\mathrm{cls}}{=}3.0$, $\lambda_{\mathrm{con}}{=}1.0$. The sensitivity to $\lambda_{\mathrm{edl}}$, the annealing step, $\lambda_{\mathrm{cls}}$, and $\lambda_{\mathrm{con}}$ is reported in Section~3.7; $\lambda_{\mathrm{cls}}^{(r)}$ and $\lambda_{\mathrm{div}}$ are held fixed across all experiments.

\begin{table*}[!t]
	\centering
	\caption{Main benchmark results across four cancer subtype classification tasks (5-fold cross-validation, mean $\pm$ std). Bold indicates the best result per task.}
	\label{tab:main_benchmark}
	\footnotesize
	\begin{tabular*}{\textwidth}{@{\extracolsep\fill}llcccccc@{\extracolsep\fill}}
		\specialrule{1.0pt}{0pt}{0pt}
		Task & Model & Accuracy & Macro-F1 & Weighted-F1 & Macro-Recall & Macro-AUC & Macro-AUPRC \\
		\specialrule{0.6pt}{0pt}{0pt}
		BRCA & CMGL & \textbf{0.8569 $\pm$ 0.0313} & \textbf{0.8184 $\pm$ 0.0577} & \textbf{0.8565 $\pm$ 0.0319} & \textbf{0.7791 $\pm$ 0.0719} & \textbf{0.9703 $\pm$ 0.0108} & \textbf{0.8756 $\pm$ 0.0579} \\
		 & HTML & 0.8360 $\pm$ 0.0185 & 0.7672 $\pm$ 0.0545 & 0.8330 $\pm$ 0.0223 & 0.6035 $\pm$ 0.0436 & 0.9500 $\pm$ 0.0172 & 0.8217 $\pm$ 0.0477 \\
		 & MOCapsNet & 0.8212 $\pm$ 0.0263 & 0.7525 $\pm$ 0.0613 & 0.8140 $\pm$ 0.0291 & 0.6364 $\pm$ 0.0914 & 0.9318 $\pm$ 0.0211 & 0.7998 $\pm$ 0.0551 \\
		 & MOGONET & 0.7556 $\pm$ 0.0245 & 0.6431 $\pm$ 0.0606 & 0.7431 $\pm$ 0.0249 & 0.6512 $\pm$ 0.1030 & 0.8963 $\pm$ 0.0299 & 0.7143 $\pm$ 0.0786 \\
		 & MOSGAT & 0.7437 $\pm$ 0.0416 & 0.6073 $\pm$ 0.0853 & 0.7371 $\pm$ 0.0469 & 0.5343 $\pm$ 0.0897 & 0.8783 $\pm$ 0.0430 & 0.6757 $\pm$ 0.0997 \\
	\cmidrule{1-8}
GBM & CMGL & \textbf{0.8238 $\pm$ 0.0871} & \textbf{0.8320 $\pm$ 0.0800} & \textbf{0.8220 $\pm$ 0.0879} & \textbf{0.8455 $\pm$ 0.0659} & \textbf{0.9653 $\pm$ 0.0242} & \textbf{0.9113 $\pm$ 0.0543} \\
		 & HTML & 0.7949 $\pm$ 0.0890 & 0.7767 $\pm$ 0.0869 & 0.7949 $\pm$ 0.0889 & 0.7635 $\pm$ 0.0632 & 0.9538 $\pm$ 0.0207 & 0.8811 $\pm$ 0.0479 \\
		 & MOCapsNet & 0.7254 $\pm$ 0.0525 & 0.7432 $\pm$ 0.0496 & 0.7243 $\pm$ 0.0547 & 0.6259 $\pm$ 0.0673 & 0.9347 $\pm$ 0.0313 & 0.8394 $\pm$ 0.0684 \\
		 & MOGONET & 0.6145 $\pm$ 0.0724 & 0.5871 $\pm$ 0.0753 & 0.5956 $\pm$ 0.0810 & 0.6340 $\pm$ 0.0684 & 0.8582 $\pm$ 0.0354 & 0.6902 $\pm$ 0.0479 \\
		 & MOSGAT & 0.6477 $\pm$ 0.0559 & 0.6422 $\pm$ 0.0486 & 0.6446 $\pm$ 0.0504 & 0.6220 $\pm$ 0.0808 & 0.8522 $\pm$ 0.0308 & 0.6971 $\pm$ 0.0543 \\
	\cmidrule{1-8}
LGG & CMGL & \textbf{0.9595 $\pm$ 0.0287} & \textbf{0.9490 $\pm$ 0.0387} & \textbf{0.9584 $\pm$ 0.0299} & \textbf{0.9391 $\pm$ 0.0325} & 0.9811 $\pm$ 0.0127 & 0.9647 $\pm$ 0.0234 \\
		 & HTML & 0.9393 $\pm$ 0.0129 & 0.9272 $\pm$ 0.0171 & 0.9387 $\pm$ 0.0130 & 0.9388 $\pm$ 0.0490 & \textbf{0.9855 $\pm$ 0.0162} & \textbf{0.9771 $\pm$ 0.0202} \\
		 & MOCapsNet & 0.9475 $\pm$ 0.0270 & 0.9382 $\pm$ 0.0329 & 0.9464 $\pm$ 0.0275 & 0.8992 $\pm$ 0.0321 & 0.9772 $\pm$ 0.0150 & 0.9591 $\pm$ 0.0233 \\
		 & MOGONET & 0.9192 $\pm$ 0.0282 & 0.9122 $\pm$ 0.0234 & 0.9174 $\pm$ 0.0273 & 0.8941 $\pm$ 0.0394 & 0.9538 $\pm$ 0.0135 & 0.9165 $\pm$ 0.0209 \\
		 & MOSGAT & 0.9191 $\pm$ 0.0529 & 0.9033 $\pm$ 0.0633 & 0.9166 $\pm$ 0.0544 & 0.8996 $\pm$ 0.0630 & 0.9692 $\pm$ 0.0255 & 0.9542 $\pm$ 0.0345 \\
	\cmidrule{1-8}
		OV & CMGL & \textbf{0.7991 $\pm$ 0.0378} & \textbf{0.7918 $\pm$ 0.0428} & \textbf{0.7955 $\pm$ 0.0420} & \textbf{0.7834 $\pm$ 0.0869} & \textbf{0.9463 $\pm$ 0.0214} & \textbf{0.8766 $\pm$ 0.0493} \\
		 & HTML & 0.7076 $\pm$ 0.0533 & 0.7083 $\pm$ 0.0573 & 0.7096 $\pm$ 0.0558 & 0.7631 $\pm$ 0.0460 & 0.9260 $\pm$ 0.0248 & 0.8369 $\pm$ 0.0506 \\
		 & MOCapsNet & 0.7568 $\pm$ 0.0764 & 0.7547 $\pm$ 0.0782 & 0.7566 $\pm$ 0.0767 & 0.6969 $\pm$ 0.0209 & 0.9382 $\pm$ 0.0315 & 0.8754 $\pm$ 0.0635 \\
		 & MOGONET & 0.6827 $\pm$ 0.0644 & 0.6768 $\pm$ 0.0696 & 0.6791 $\pm$ 0.0662 & 0.6491 $\pm$ 0.0676 & 0.8591 $\pm$ 0.0407 & 0.7242 $\pm$ 0.0776 \\
		 & MOSGAT & 0.6019 $\pm$ 0.0309 & 0.5926 $\pm$ 0.0321 & 0.5955 $\pm$ 0.0287 & 0.6025 $\pm$ 0.0549 & 0.8507 $\pm$ 0.0391 & 0.6930 $\pm$ 0.0841 \\
		\specialrule{1.0pt}{0pt}{0pt}
	\end{tabular*}
	\par\medskip
	\caption{Pan-cancer benchmark results (5-fold cross-validation, mean $\pm$ std). Bold indicates the best result.}
	\label{tab:pancancer_full}
	\normalsize
	\begin{tabular*}{\textwidth}{@{\extracolsep\fill}lcccccc@{\extracolsep\fill}}
		\specialrule{1.0pt}{0pt}{0pt}
		Model & Accuracy & Macro-F1 & Weighted-F1 & Macro-Recall & Macro-AUC & Macro-AUPRC \\
		\specialrule{0.6pt}{0pt}{0pt}
		CMGL & $\mathbf{0.9682 \pm 0.0018}$ & $\mathbf{0.9483 \pm 0.0047}$ & $\mathbf{0.9676 \pm 0.0022}$ & $\mathbf{0.9448 \pm 0.0148}$ & $\mathbf{0.9984 \pm 0.0011}$ & $\mathbf{0.9746 \pm 0.0047}$ \\
		HTML & $0.9577 \pm 0.0028$ & $0.9375 \pm 0.0064$ & $0.9580 \pm 0.0032$ & $0.8822 \pm 0.0084$ & $0.9983 \pm 0.0003$ & $0.9568 \pm 0.0069$ \\
		MOCapsNet & $0.9464 \pm 0.0096$ & $0.9054 \pm 0.0146$ & $0.9399 \pm 0.0104$ & $0.8004 \pm 0.0117$ & $0.9976 \pm 0.0011$ & $0.9619 \pm 0.0104$ \\
		MOGONET & $0.9666 \pm 0.0034$ & $0.9364 \pm 0.0199$ & $0.9649 \pm 0.0051$ & $0.9222 \pm 0.0154$ & $0.9981 \pm 0.0013$ & $0.9701 \pm 0.0066$ \\
		MOSGAT & $0.9643 \pm 0.0030$ & $0.9050 \pm 0.0094$ & $0.9623 \pm 0.0036$ & $0.9010 \pm 0.0111$ & $0.9976 \pm 0.0014$ & $0.9566 \pm 0.0067$ \\
		\specialrule{1.0pt}{0pt}{0pt}
	\end{tabular*}
\end{table*}

\subsection{Main Benchmark Results}

Table~\ref{tab:main_benchmark} presents the 5-fold results on the four single-cancer tasks; a bar-chart view is in Supplementary Fig.~\ref{fig:benchmark_bar}. CMGL attains the highest accuracy and Macro-F1 on all four tasks. Its average accuracy is 0.8598 and its average Macro-F1 is 0.8478, surpassing the second-best method HTML by 4.03\% in accuracy. The advantage of independent confidence estimation is most apparent on tasks where modality quality varies substantially: on GBM, whose 244 samples carry noisy per-modality signal, CMGL leads HTML by 2.89\% \rev{($p_{\mathrm{adj}}=0.0294$)} in accuracy and 5.53\% \rev{($p_{\mathrm{adj}}=0.0091$)} in Macro-F1, matching the expectation that explicit reliability priors are more useful when the signal to noise ratio is low. The smaller margin on LGG, where subtype boundaries are already separable from individual modalities, further supports this interpretation. \rev{Additional paired tests show that CMGL significantly outperforms the baselines in most cancer subtype classification comparisons (Supplementary Table~\ref{tab:cover_all_baseline_fold_tests}).}

\subsection{Pan-cancer Benchmark}

The 32-class pan-cancer task tests whether CMGL scales to a setting where each cancer type has distinct omics profiles and class boundaries overlap in complex ways (Table~\ref{tab:pancancer_full}; a radar-chart visualization of the same data is given in Supplementary Fig.~\ref{fig:pancancer_radar}). CMGL achieves the best performance across all six metrics (accuracy 0.9682, Macro-F1 0.9483, Macro-Recall 0.9448). \rev{Its Macro-F1 exceeds that of MOGONET ($p_{\mathrm{adj}}=0.0002$).} The smaller margin relative to the single-cancer tasks is unsurprising: pan-cancer classification already has strong per-modality discriminative signal, so confidence gating is most useful for within-cancer tasks.

\subsection{Small-sample Generalization}

\begin{figure*}[!t]
	\centering
	\includegraphics[width=\textwidth]{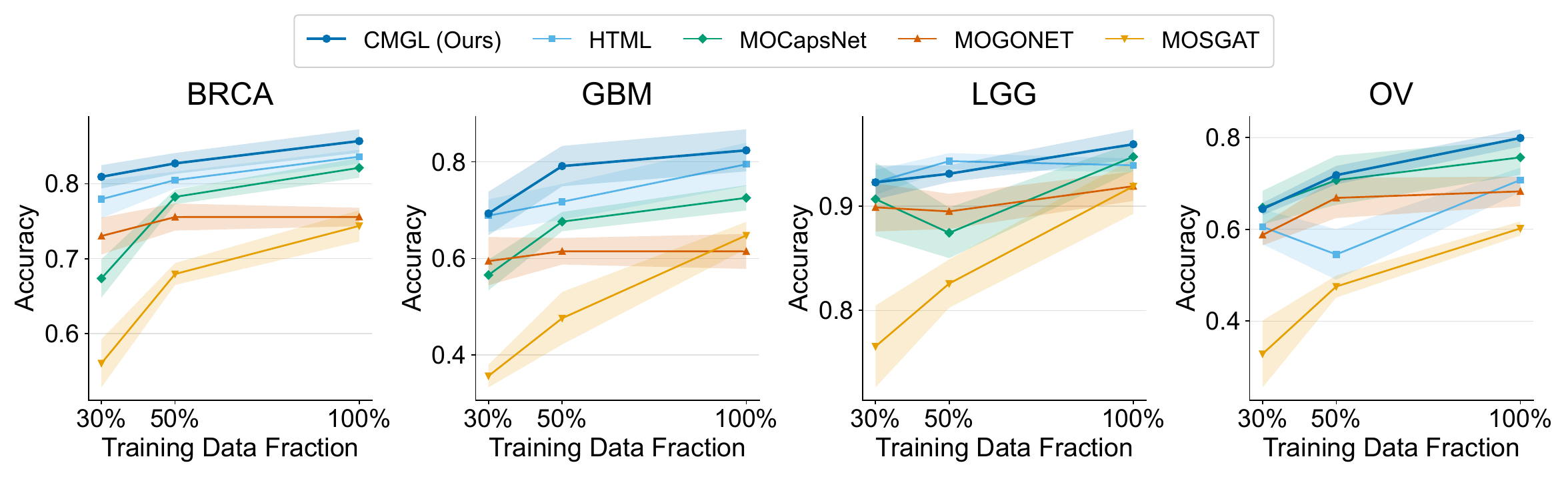}
	\caption{Small-sample generalization curves across the four cancer subtype classification tasks.}
	\label{fig:smallsample}
\figalttext[Small-sample performance across four cancer tasks]{Alt text: Four line plots compare Accuracy at 30, 50, and 100 percent training data for five methods on BRCA, GBM, LGG, and OV; CMGL is strongest overall, particularly on BRCA and GBM.}%
\end{figure*}

Because clinical cohorts for rare cancer subtypes are often small, we assess whether the confidence-guided mechanism of CMGL remains effective under limited supervision by evaluating performance at training fractions of 0.3 and 0.5, with the full-data setting as reference (Fig.~\ref{fig:smallsample}). CMGL delivers the strongest overall small-sample performance across the four tasks, with the clearest advantage on BRCA and GBM as the training set shrinks. Although HTML or MOCapsNet lead in a few isolated settings, CMGL remains the most consistently strong method across tasks and fractions, indicating that frozen confidence priors continue to provide an advantage when labels are scarce even though absolute accuracy still drops at lower training fractions. Per-task numerical results are provided in Supplementary Table~\ref{tab:smallsample_full}.

\begin{figure*}[!t]
\centering
\includegraphics[width=\textwidth]{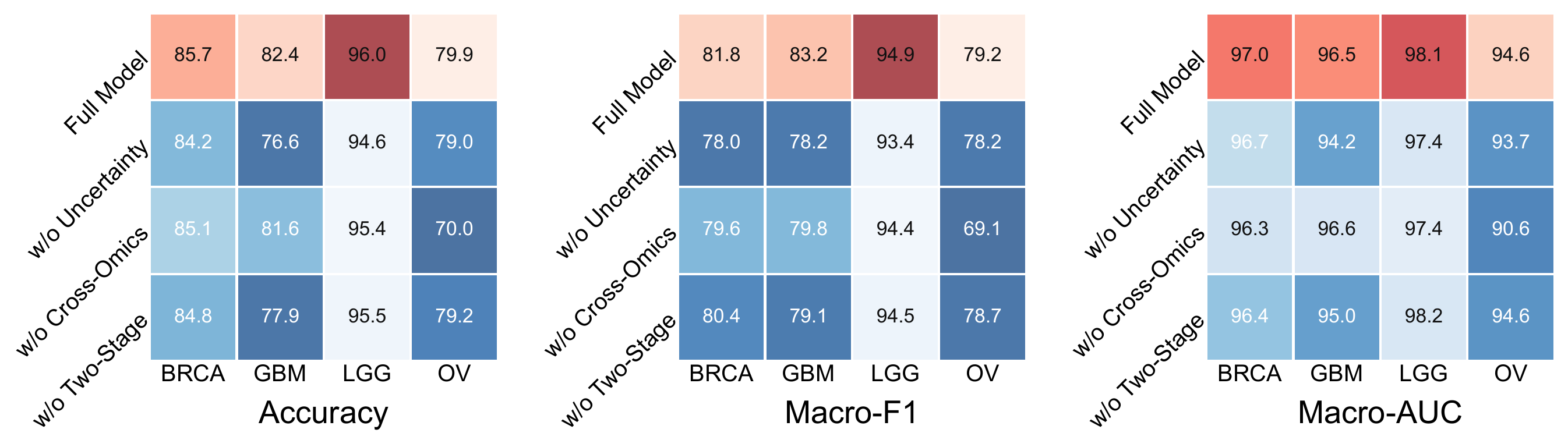}
\caption{Ablation heatmap of Accuracy, Macro-F1, and Macro-AUC across the four benchmark tasks.}
\label{fig:ablation}
\figalttext[CMGL component ablation heatmaps]{Alt text: Three heatmaps compare the full model with three component removals across four tasks; each removal lowers at least one metric, with cross-omics fusion having the largest average effect.}%
\end{figure*}

\vspace{2\baselineskip}
\subsection{\revheading{Modality Confidence Analysis}}

\rev{To assess whether Stage~1 confidence tracks prediction sensitivity, we removed each modality during frozen checkpoint inference. We then compared its confidence with the resulting changes in true class probability and negative log likelihood across 15 matched evaluations from three seeds and five folds (Table~\ref{tab:confidence_occlusion_main}).}

\begin{table}[H]
\centering
\caption{\rev{Patient level agreement between Stage~1 confidence and classification impact after modality removal. The two $\rho$ values are Spearman rank correlations between modality confidence and the change in true class probability ($\Delta p_{\rm true}$) or negative log likelihood ($\Delta\mathrm{NLL}$).}}
\label{tab:confidence_occlusion_main}
\scriptsize
\begin{tabular*}{\columnwidth}{@{\extracolsep\fill}lccc@{\extracolsep\fill}}
\specialrule{1.0pt}{0pt}{0pt}
Task & $\rho$(conf., $\Delta p_{\rm true}$) & $\rho$(conf., $\Delta\mathrm{NLL}$) & Top-1 agreement \\
\specialrule{0.6pt}{0pt}{0pt}
BRCA & $0.2639\pm0.2085$ & $0.2574\pm0.2027$ & $0.7074\pm0.2017$ \\
GBM & $0.3573\pm0.1292$ & $0.3404\pm0.1409$ & $0.7310\pm0.1724$ \\
LGG & $0.4523\pm0.1000$ & $0.4479\pm0.0993$ & $0.8326\pm0.0925$ \\
OV & $0.2362\pm0.1781$ & $0.2080\pm0.1938$ & $0.6386\pm0.2113$ \\
\specialrule{1.0pt}{0pt}{0pt}
\end{tabular*}
\end{table}

\rev{The learned confidence scores quantify the relative predictive relevance of each available modality for an individual patient. Both associations are positive on all four tasks, with the strongest agreement on LGG ($\rho=0.4523$ for $\Delta p_{\rm true}$ and $0.4479$ for $\Delta\mathrm{NLL}$). Top-1 agreement ranges from 0.6386 on OV to 0.8326 on LGG, showing that the modality assigned the highest confidence usually has the largest effect on the same patient's prediction when removed. The positive associations and Top-1 agreement show that Stage~1 provides a useful signal for confidence guided fusion at the patient level. Classification changes by modality and results for each seed are reported in Supplementary Tables~\ref{tab:confidence_occlusion_detail} and~\ref{tab:cover_confidence_detailed}.}

\subsection{Ablation Study}

To test whether reliability estimation, cross-omics fusion, and two-stage separation account for the benchmark gains, we remove each component in turn under the same evaluation protocol (Fig.~\ref{fig:ablation}; full numerical table in Supplementary Table~\ref{tab:ablation_full}). Cross-omics attention is the most impactful module: removing it costs 3.0\% in average accuracy and 4.1\% in Macro-F1, and as much as 9.9\% in accuracy on OV. OV is the task with the most heterogeneous signal strengths across modalities, so feature-wise exchange matters most where no single modality dominates; naive concatenation or averaging loses this complementarity. Removing uncertainty-aware learning reduces the average accuracy by 2.4\%, with the largest drop on GBM (5.8\%). GBM is the smallest dataset with the lowest signal to noise ratio, the regime in which explicit confidence estimation compensates for the limited classification signal. Two-stage decoupling has a smaller effect ($-1.6$\%) but again matters most on GBM ($-4.5$\%) and least on LGG, where subtype boundaries are well separated without reliability priors.

\rev{To determine whether softer graph topologies improve the confidence guided framework, we compared the default intersection graph with union and confidence-topk under equivalent experimental conditions. Intersection provided the strongest overall classification result and remained the sparsest strategy on every task. This finding shows that confidence guided fusion and consensus graph construction serve complementary roles, with confidence regulating modality contributions and agreement across views limiting the propagation of structural noise from individual views (Supplementary Table~\ref{tab:graph_strategy_main}).}

\rev{Because multi-omics measurements are often incomplete in clinical datasets, we evaluated CMGL under partial modality availability at the cohort level using all distinct combinations of two views and three views, with the full model using four views as the reference under equivalent experimental conditions. For each subset of available modalities, CMGL was retrained from initialization using only those views: Stage~1 learned confidence scores from the available omics, separate $k$-NN graphs were constructed from the same inputs, their intersection defined the patient graph, and Stage~2 used the subset confidence scores to fuse representations and train the GraphSAGE classifier under the same experimental settings. Average Accuracy and Macro-F1 increased with broader modality coverage on all four tasks. The complete summary and results for each combination appear in Supplementary Tables~\ref{tab:modality_ablation_main} and~\ref{tab:cover_all_modality_combinations}.}

\setcounter{figure}{4}%
	\begin{figure*}[!t]
		\centering
		\begin{minipage}[t]{0.48\textwidth}
			\centering
			\includegraphics[width=\textwidth]{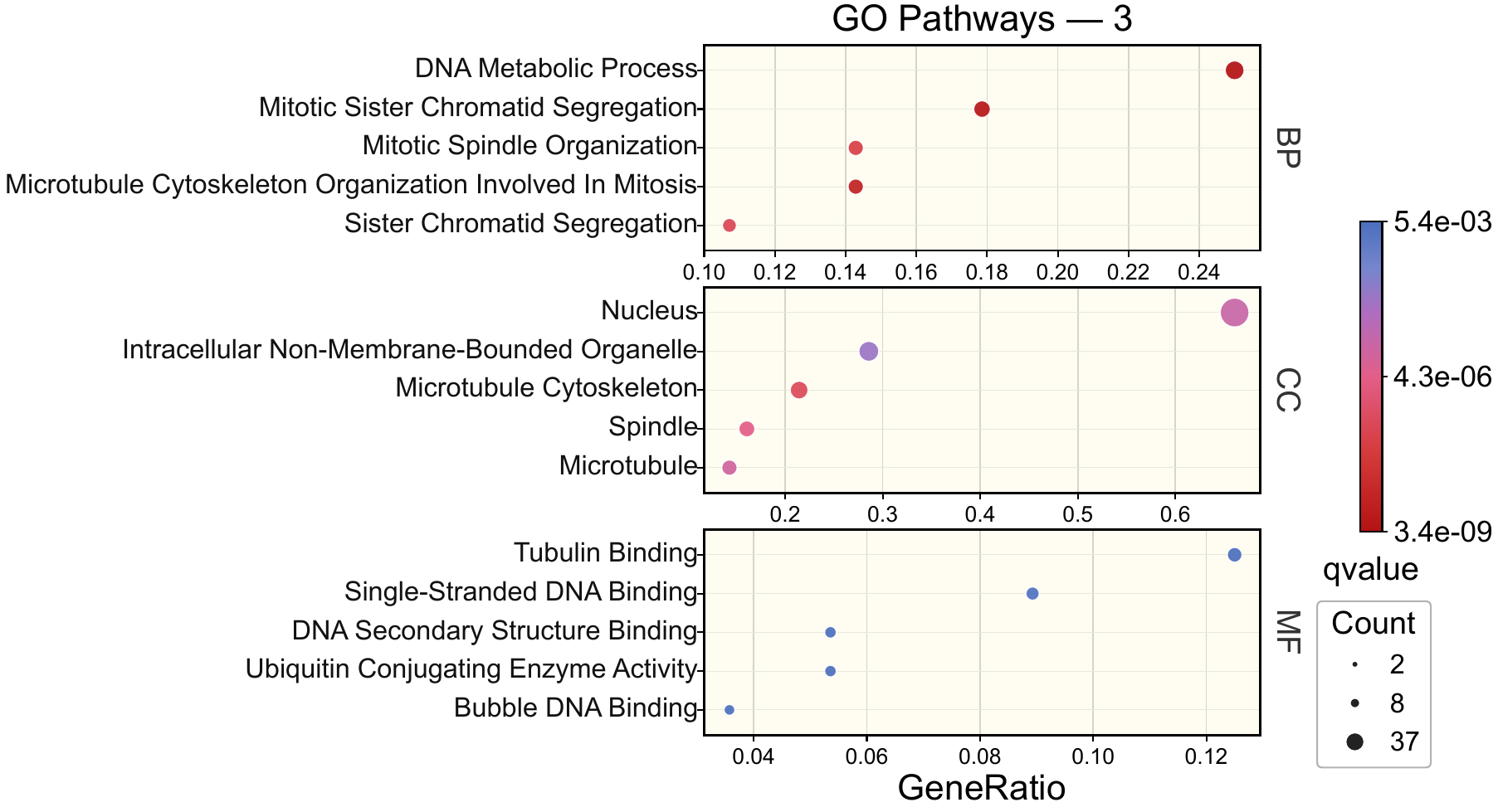}\\[2pt]
			{\footnotesize Class 2 (Basal-like)}
		\end{minipage}\hfill
		\begin{minipage}[t]{0.48\textwidth}
			\centering
			\includegraphics[width=\textwidth]{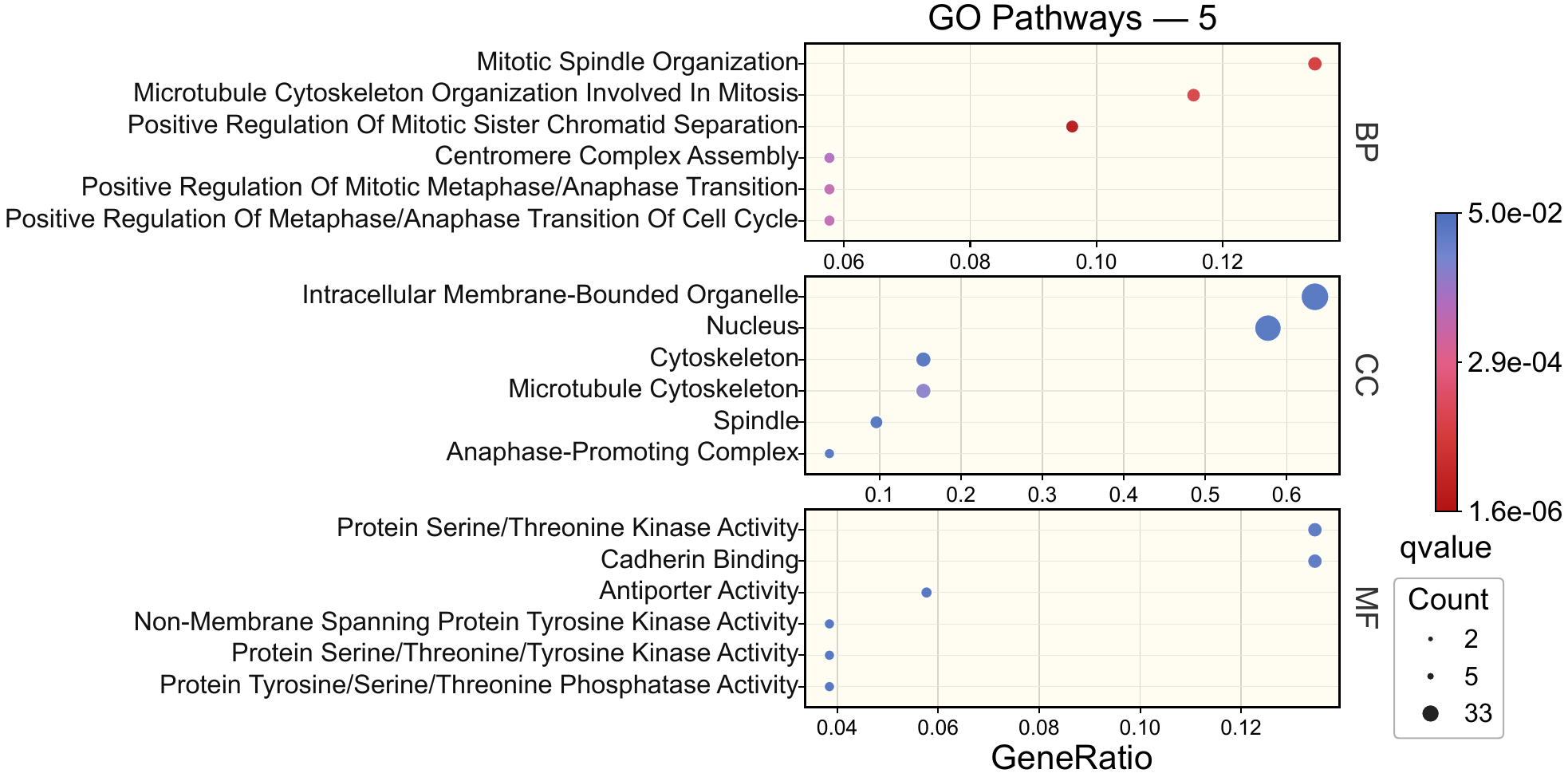}\\[2pt]
			{\footnotesize Class 4 (Luminal~B)}
		\end{minipage}
		\caption{Representative GO Biological Process enrichment plots for the two most proliferation-associated BRCA classes (Class~2, Basal-like; Class~4, Luminal~B). Enrichment plots for all five classes are provided in Supplementary Fig.~\ref{fig:brca_go_all}.}
		\label{fig:brca_go}
	\figalttext[GO enrichment for BRCA Classes 2 and 4]{Alt text: Two bubble plots show GO enrichment for BRCA Classes 2 and 4; both profiles are dominated by cell-cycle, mitotic, and microtubule-related processes.}%
	\end{figure*}

\begin{table*}[!t]
	\centering
		\caption{Summary of the five predicted BRCA subtypes: sample counts, top marker genes, representative enriched pathways, and putative PAM50 correspondence.}
	\label{tab:brca_subtypes}
	\footnotesize
	\begin{tabular*}{\textwidth}{@{\extracolsep\fill}cclllc@{\extracolsep\fill}}
		\specialrule{1.0pt}{0pt}{0pt}
		Class & $n$ (\%) & Top Marker Gene & Representative Pathway (adj.\ $p$) & Molecular Feature & PAM50 \\
		\specialrule{0.6pt}{0pt}{0pt}
		0 & 366 (54.5\%) & ESR1, MAPT & Low proliferation (CENPN$\downarrow$, CDC20$\downarrow$) & HR$+$, low-prolif. & Luminal~A \\
		1 & 40 (6.0\%) & MPHOSPH6 & Trp-kynurenine catabolism ($1.3\!\times\!10^{-4}$) & Metabolic/immunomod. & Non-PAM50 \\
		2 & 121 (18.0\%) & EGFR & DNA Metabolic Process ($3.4\!\times\!10^{-9}$) & High-prolif., cell cycle & Basal-like \\
		3 & 31 (4.6\%) & ATOH8 & Aorta Development ($2.6\!\times\!10^{-2}$) & Mesenchymal/develop. & Normal-like \\
		4 & 113 (16.8\%) & MLPH & Mitotic Spindle Org. ($1.5\!\times\!10^{-5}$) & Luminal marker MLPH & Luminal~B \\
		\specialrule{1.0pt}{0pt}{0pt}
	\end{tabular*}
\end{table*}

\newcommand{\kircKmFigure}{%
\begin{figure*}[!t]
\centering
\begin{minipage}[t]{0.32\textwidth}
\centering
\includegraphics[width=\textwidth]{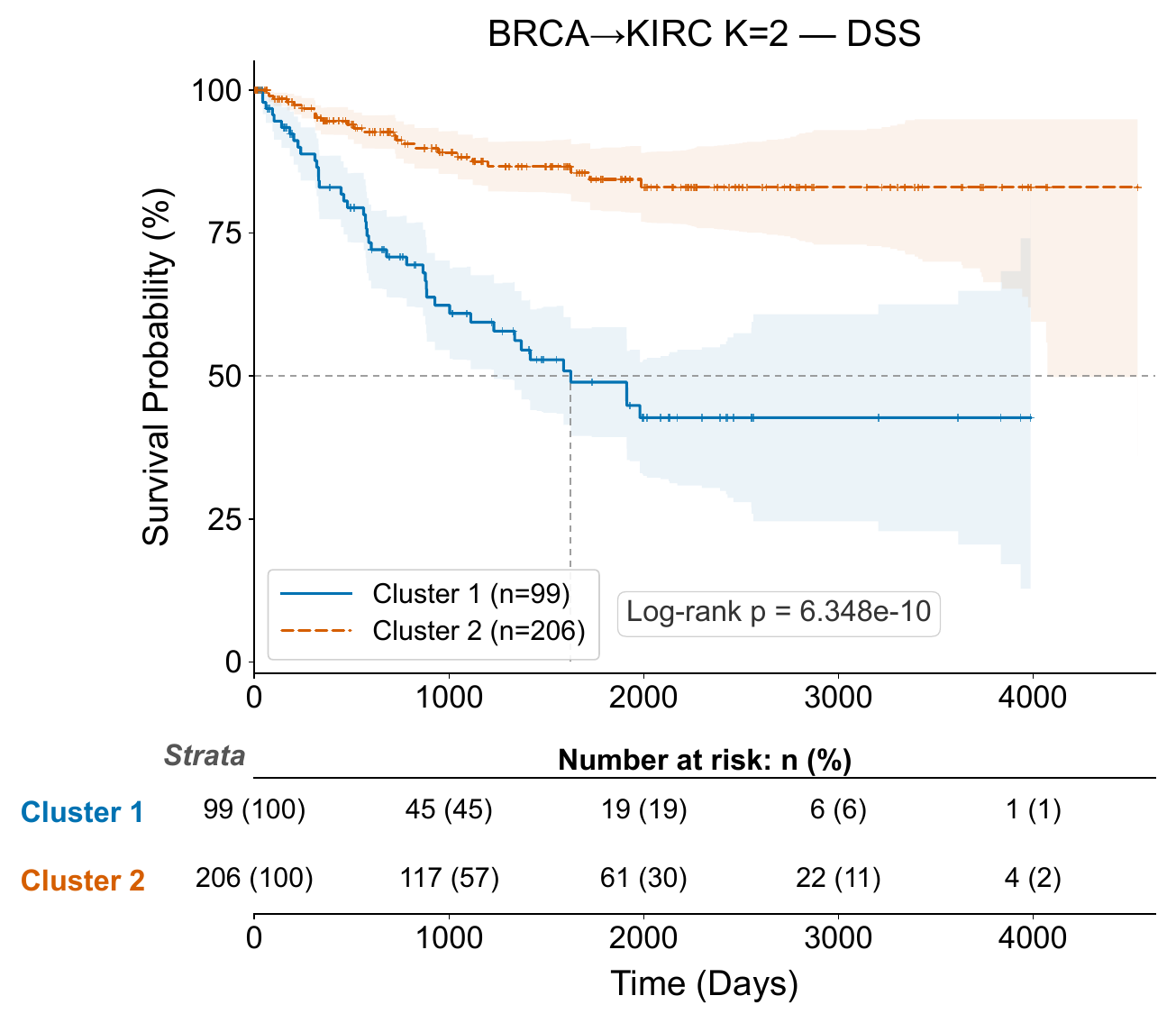}\\[2pt]
{\footnotesize (a) DSS, $K=2$}
\end{minipage}\hfill
\begin{minipage}[t]{0.32\textwidth}
\centering
\includegraphics[width=\textwidth]{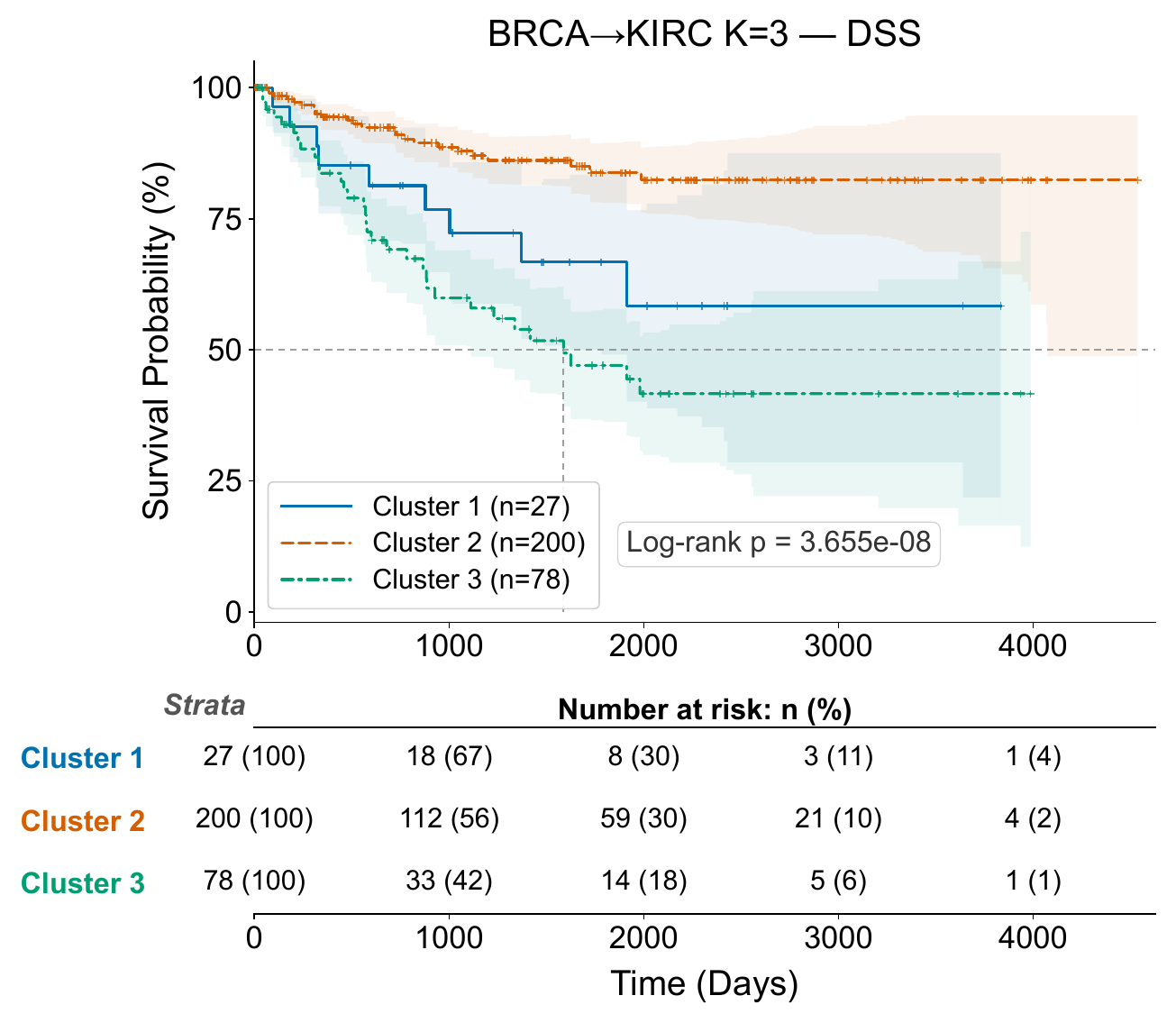}\\[2pt]
{\footnotesize (b) DSS, $K=3$}
\end{minipage}\hfill
\begin{minipage}[t]{0.32\textwidth}
\centering
\includegraphics[width=\textwidth]{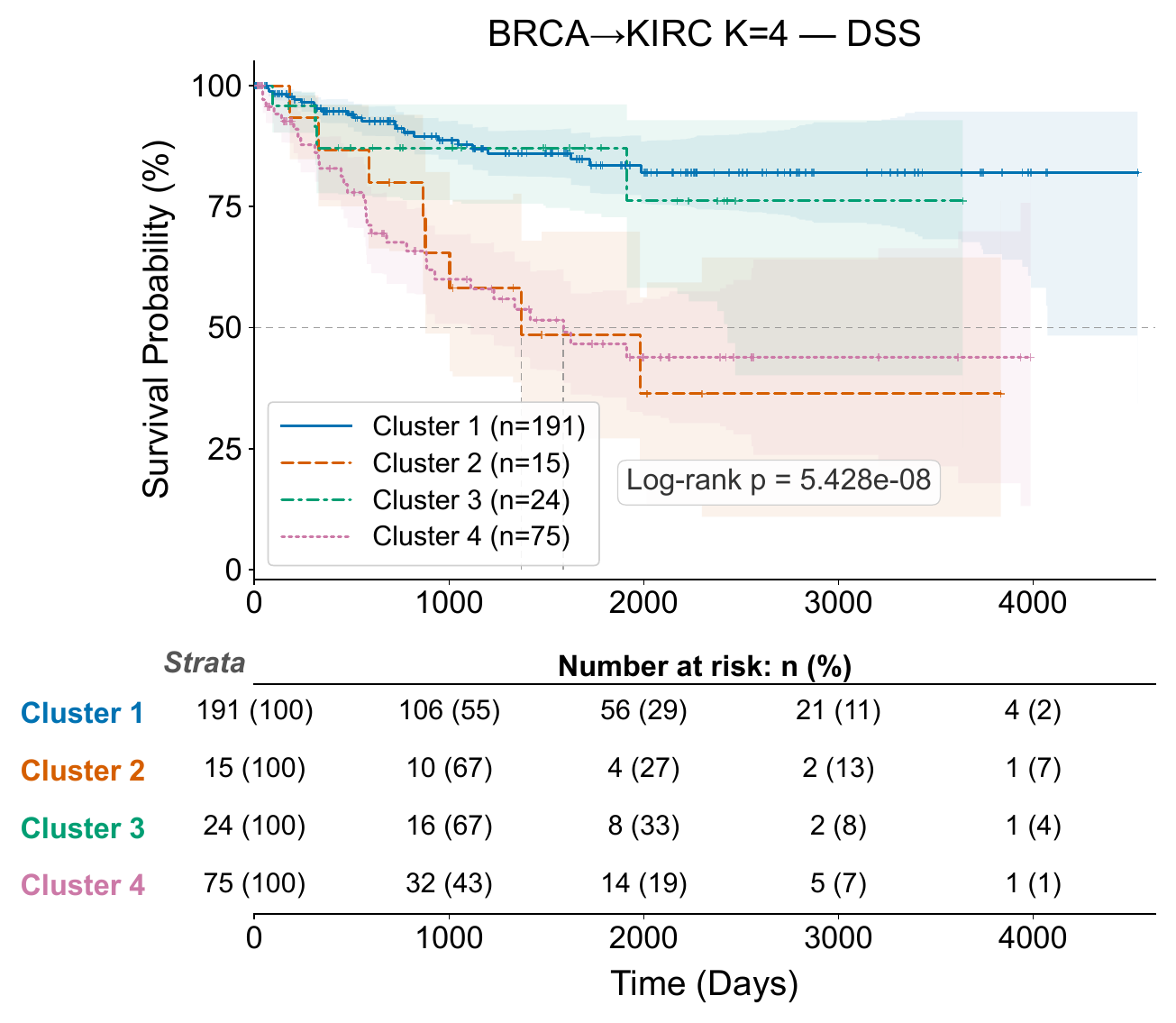}\\[2pt]
{\footnotesize (c) DSS, $K=4$}
\end{minipage}\\[6pt]
\begin{minipage}[t]{0.32\textwidth}
\centering
\includegraphics[width=\textwidth]{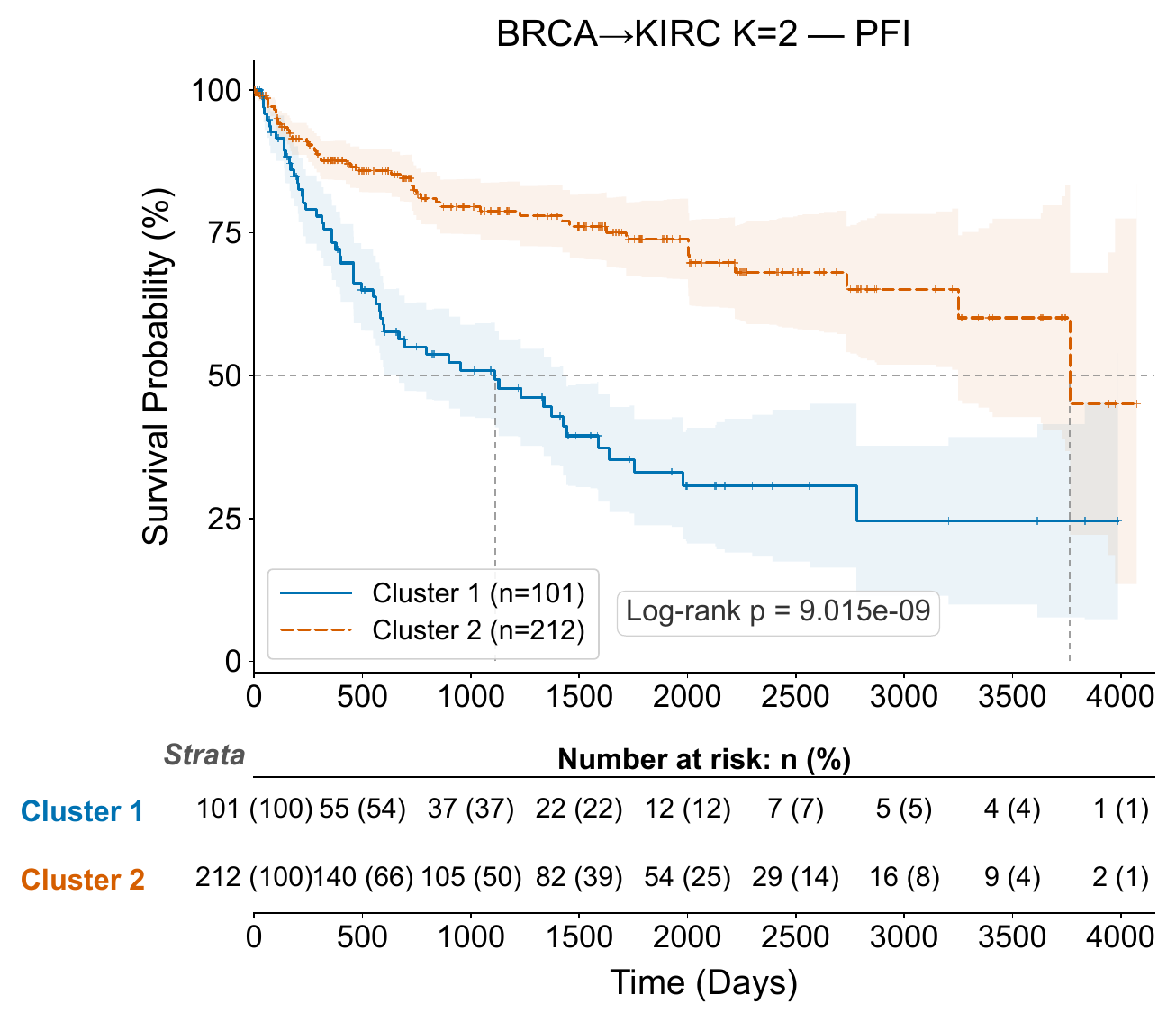}\\[2pt]
{\footnotesize (d) PFI, $K=2$}
\end{minipage}\hfill
\begin{minipage}[t]{0.32\textwidth}
\centering
\includegraphics[width=\textwidth]{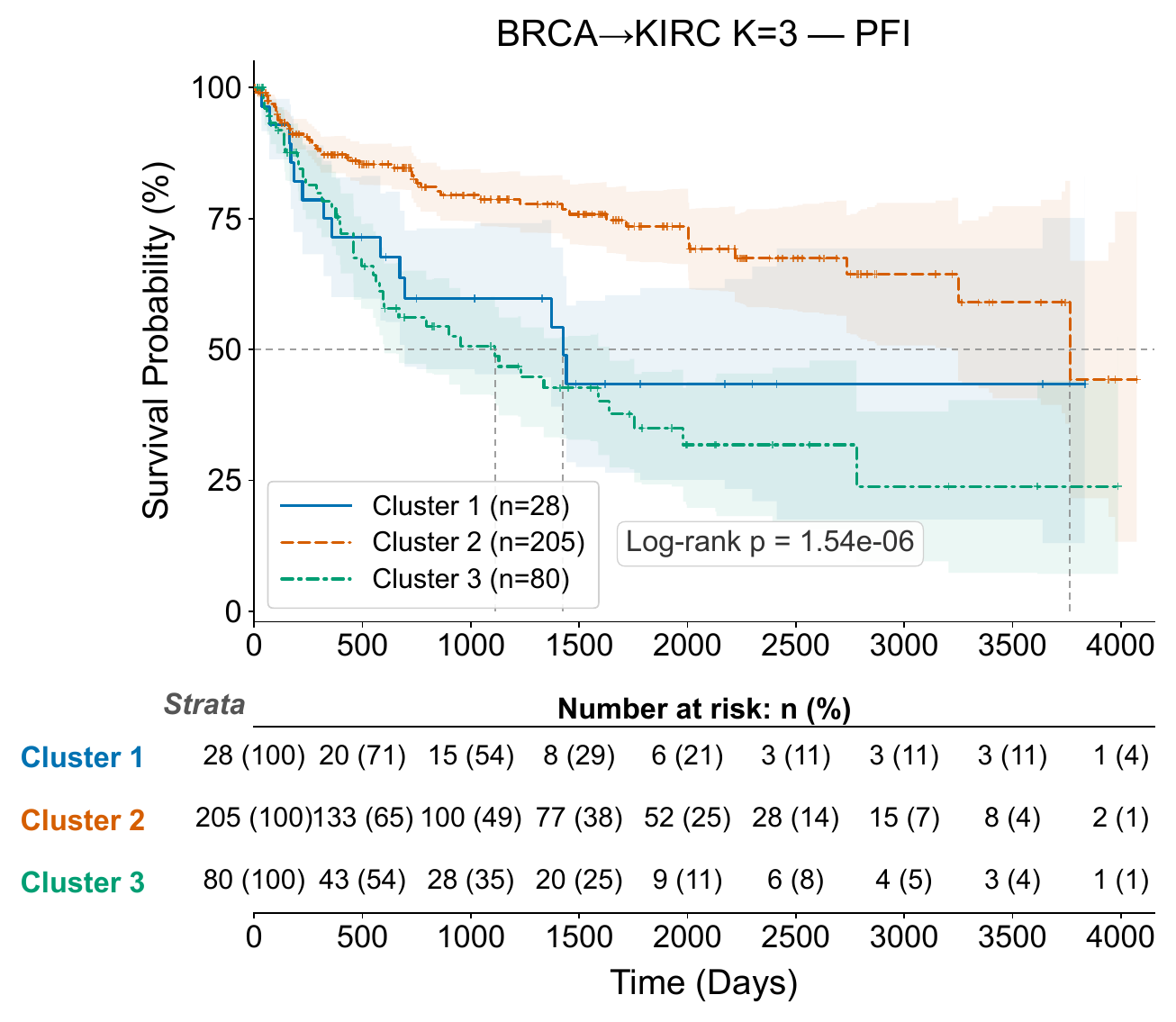}\\[2pt]
{\footnotesize (e) PFI, $K=3$}
\end{minipage}\hfill
\begin{minipage}[t]{0.32\textwidth}
\centering
\includegraphics[width=\textwidth]{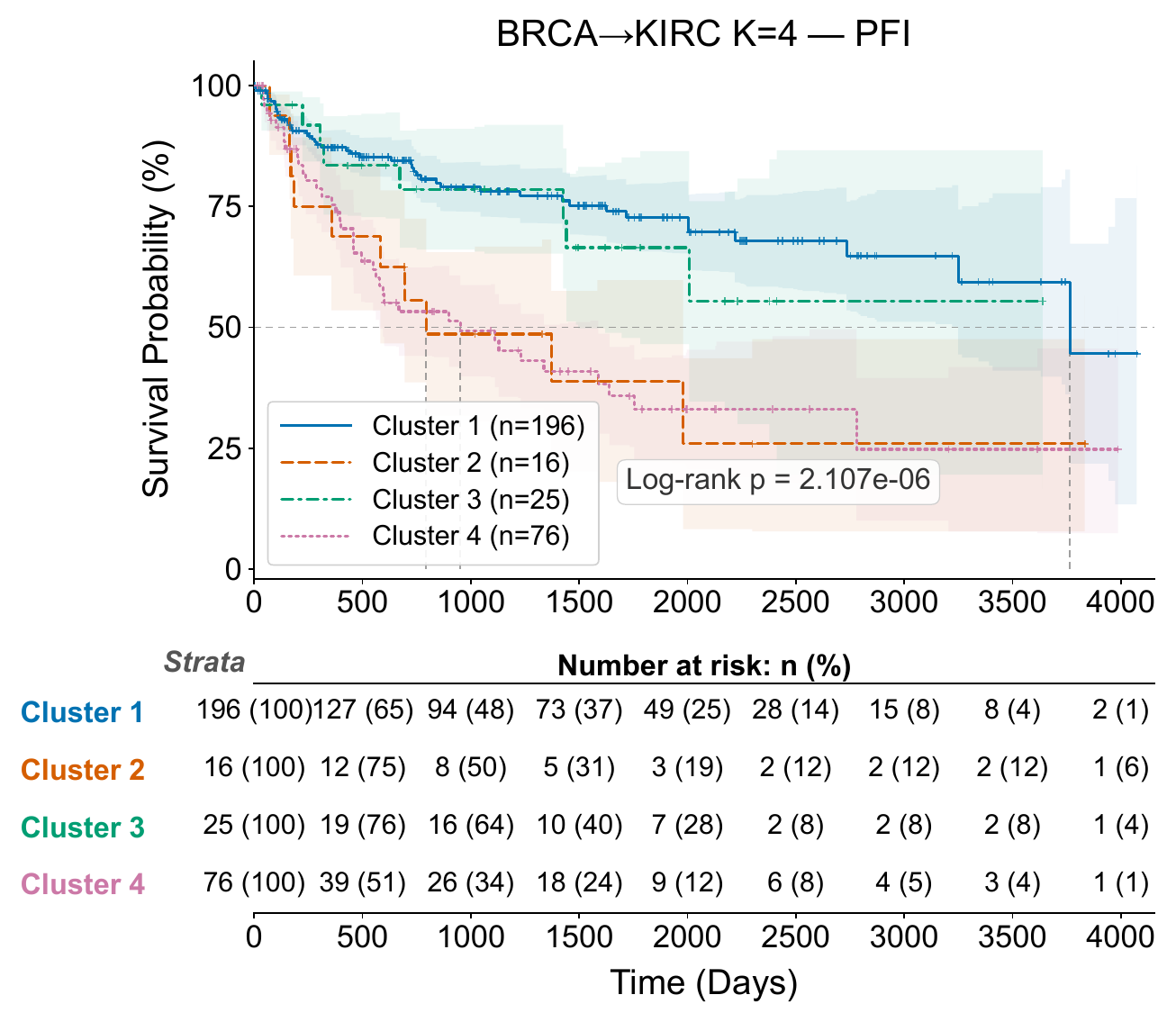}\\[2pt]
{\footnotesize (f) PFI, $K=4$}
\end{minipage}
\caption{KIRC Kaplan-Meier curves under the BRCA-transferred embedding space. Top row: disease-specific survival (DSS); bottom row: progression-free interval (PFI). Columns correspond to $K=2$, $K=3$, and $K=4$ clustering solutions.}
\label{fig:kirc_km}
\figalttext[KIRC survival curves across three clustering resolutions]{Alt text: Six Kaplan-Meier panels compare disease-specific survival and progression-free interval among KIRC clusters at K equals 2, 3, and 4; proliferation-associated clusters consistently have poorer survival.}%
\end{figure*}
}

\setcounter{figure}{5}%
\kircKmFigure

\subsection{Hyperparameter Sensitivity Analysis}

We next examine the sensitivity of CMGL to its loss-weight hyperparameters on BRCA under the same 5-fold protocol. Four coefficients are jointly varied on two disjoint $3\!\times\!3$ grids: a Stage-1 sub-grid over $\lambda_{\mathrm{edl}}$ and the EDL annealing step, and a Stage-2 sub-grid over $\lambda_{\mathrm{cls}}$ and $\lambda_{\mathrm{con}}$. Across all 18 configurations, Macro-F1 varies by less than $\pm 0.025$ (Fig.~\ref{fig:hparam_3dcolumn}; full grid in Supplementary Table~\ref{tab:hparam_grid}), so CMGL is robust to these loss weights within the swept ranges.

The $k$-NN neighbor count $k$ controls both graph density and the effective smoothing radius of message passing, and is the one hyperparameter that shows visible per-fold sensitivity. Rather than fixing $k$ in advance, CMGL performs adaptive $k$ selection via the warm up procedure in Section~2.4: for each candidate $k\in\mathcal{K}=\{7,11,15,19,23\}$ we score validation Macro-F1 and retain $k^\ast = \arg\max_k \mathrm{F1}_{\mathrm{val}}(k)$.

\setcounter{figure}{3}%
\begin{figure}[H]
	\centering
	\includegraphics[width=0.98\columnwidth]{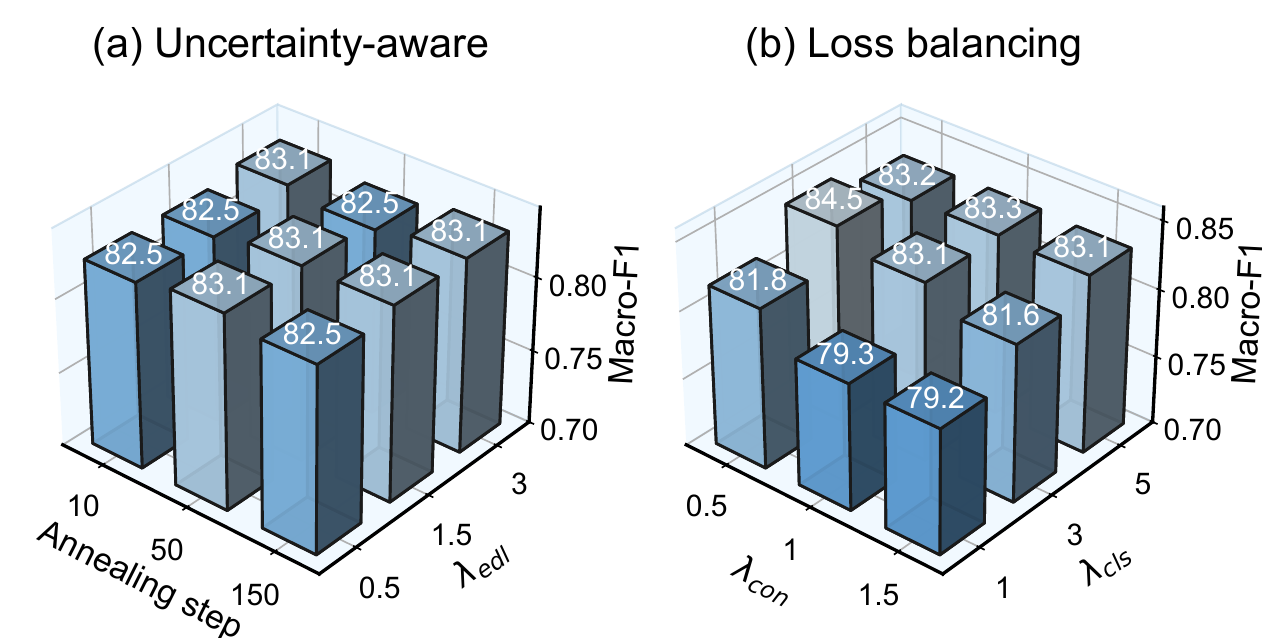}
\caption{BRCA hyperparameter sensitivity across the Stage~1 ($\lambda_{\mathrm{edl}}$, annealing step) and Stage~2 ($\lambda_{\mathrm{cls}}$, $\lambda_{\mathrm{con}}$) grids.}
	\label{fig:hparam_3dcolumn}
\figalttext[BRCA hyperparameter sensitivity]{Alt text: Two three-dimensional bar charts show BRCA Macro-F1 across the Stage 1 and Stage 2 loss-weight grids; the values form a broad performance plateau over the tested settings.}%
\end{figure}
\setcounter{figure}{6}%

\subsection{Biological Interpretability of BRCA Subtype Classification}
\label{sec:brca_interpret}

\rev{We next examine whether the representations underlying these predictive results also capture established molecular structure in BRCA.} \rev{Following the frozen inference protocol in Section~3.1, we obtain the predicted classes, embeddings from the penultimate layer, and confidence scores for each modality for all 671 samples. We then rank the top 200 mRNA marker genes per class by effect size, perform GO Biological Process and KEGG pathway enrichment, and relate the predicted classes to TCGA-CDR clinical endpoints \citep{Chen2013enrichr,Kuleshov2016}.} The silhouette coefficient of the predicted classes reaches 0.867, a value consistent with clear class separation \citep{Rousseeuw1987} (Supplementary Fig.~\ref{fig:brca_tsne}; t-SNE projection). The molecular signatures of the five predicted classes correspond closely to the established PAM50 intrinsic subtypes of breast cancer (Table~\ref{tab:brca_subtypes}); they are listed below in order of sample proportion:

\begin{itemize}
    \item \textbf{Class~0 (54.5\% of samples)}. \rev{Top markers ESR1 and MAPT, together with downregulated mitotic markers (CENPN, ORC1, CDC20, and CCNB2), define a low proliferation, hormone receptor positive profile consistent with Luminal~A.} \rev{Its sample proportion matches the known distribution of Luminal~A in the TCGA cohort for BRCA \citep{TCGA2012brca}.}
    \item \textbf{Class~2 (18.0\%)}. Top marker EGFR, with pathway enrichment heavily concentrated in DNA metabolism and cell cycle (adj.\ $p$ reaching $10^{-9}$), matching the high-proliferation Basal-like subtype.
    \item \textbf{Class~4 (16.8\%)}. Top marker MLPH (melanophilin), a known Luminal~B marker \citep{Prat2015}.
    \item \textbf{Class~1 (6.0\%)}. Top marker MPHOSPH6, with strong enrichment in tryptophan--kynurenine catabolism (adj.\ $p=1.3\times10^{-4}$) and glutathione metabolism (KEGG adj.\ $p=9.9\times10^{-4}$). \rev{The kynurenine pathway has been recognized as a key regulatory mechanism of immune suppression in the tumor microenvironment \citep{Platten2019}, suggesting that this subtype may represent a metabolically active immunomodulatory phenotype.}
    \item \textbf{Class~3 (4.6\%)}. Top marker ATOH8, with enrichment in the Aorta Development pathway (adj.\ $p=2.6\times10^{-2}$, borderline) and broader mesenchymal/developmental programs, putatively corresponding to Normal-like.
\end{itemize}

Representative GO enrichment results for the five predicted classes are shown in Fig.~\ref{fig:brca_go}, the mRNA marker heatmap is provided in Supplementary Fig.~\ref{fig:brca_heatmap}, and full per-class GO BP and KEGG enrichment terms are tabulated in Supplementary Table~\ref{tab:brca_pathway_extended}.

\FloatBarrier

\begin{figure}[H]
\centering
\includegraphics[width=0.82\columnwidth]{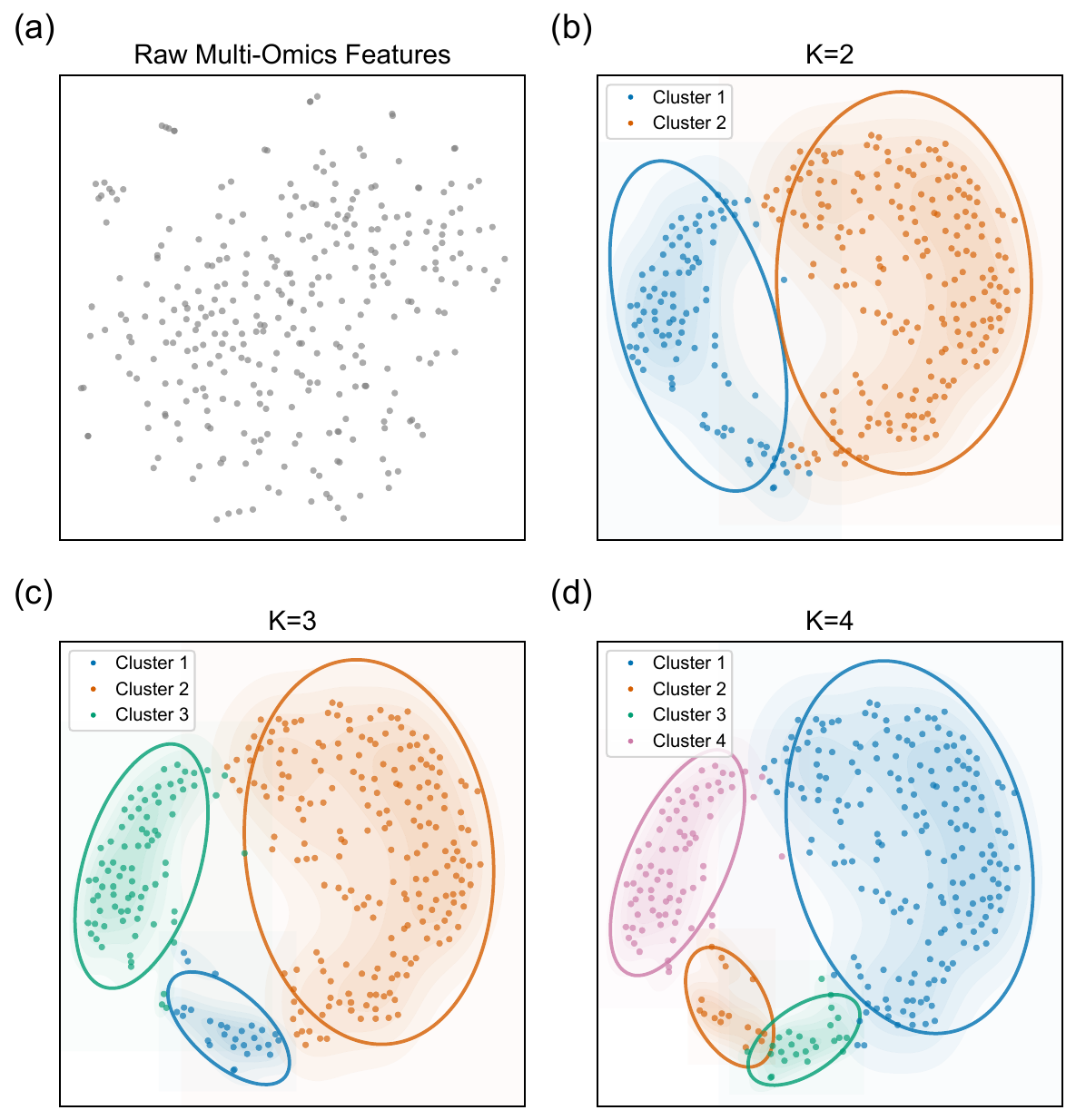}
\caption{t-SNE visualization of KIRC embeddings under the BRCA-transferred representation and KMeans clustering solutions ($K=2,3,4$).}
\label{fig:kirc_tsne}
\figalttext[KIRC embedding and clustering visualization]{Alt text: Four t-SNE panels compare raw KIRC features with CMGL embeddings clustered at K equals 2, 3, and 4; the learned embeddings form separated groups that subdivide as K increases.}%
\end{figure}

\subsection{Cross-cancer Transfer Prognostic Stratification}
\label{sec:kirc_transfer}

To test whether the CMGL embeddings transfer to other cancer types, we apply the model trained on BRCA directly to the KIRC cohort ($n=314$) without any fine tuning in the target domain and extract 64 dimensional embeddings from the penultimate layer in a single forward pass. We then run unsupervised KMeans clustering, followed by survival analysis and pathway enrichment, to see whether the proliferation and differentiation axis learned from breast cancer also stratifies renal cancer prognosis.

\subsubsection{Clustering Analysis}
\label{sec:kirc_clustering}

\rev{We evaluated KMeans solutions $K\in\{2,3,4\}$ on the frozen KIRC embeddings and selected $K=3$ using a joint criterion based on the silhouette coefficient and log rank $p$ value, with a silhouette coefficient of 0.585.} \rev{The three clusters contain 28, 206, and 80 samples. Cluster~2 has the lowest risk, is 98.1\% Luminal~A like, and corresponds mainly to source domain Class~0. Cluster~3 has the highest risk, is 88.8\% Basal like, and corresponds mainly to source domain Class~2. Cluster~2 also has the highest mean confidence (Supplementary Table~\ref{tab:kirc_clusters_all}).}

\subsubsection{Survival Analysis}
\label{sec:kirc_survival}

\newcommand{\kircPathwayTable}{%
\begin{table*}[!t]
	\centering
	\caption{Pathway enrichment analysis for KIRC clusters under $K=2,3,4$ clustering solutions. For each cluster, top-ranked terms from GO Biological Process (BP), GO Cellular Component (CC), GO Molecular Function (MF), and KEGG are shown. Bold adjusted $p$-values indicate significance at $\alpha=0.05$. N/A denotes no enriched terms returned by the library for that cluster.}
	\label{tab:kirc_pathways}
	\footnotesize
	\begin{tabular*}{\textwidth}{@{\extracolsep\fill}ccllcl@{\extracolsep\fill}}
		\specialrule{1.0pt}{0pt}{0pt}
		$K$ & Cluster ($n$) & Source & Enriched Term & adj.\ $p$ & Representative Genes \\
		\specialrule{0.6pt}{0pt}{0pt}
		\multirow{8}{*}{2} & \multirow{4}{*}{1 (101)} & GO BP & Microtubule Cytoskeleton Org.\ in Mitosis & $\mathbf{2.7\!\times\!10^{-4}}$ & CDK1, BIRC5, KIF11, NDC80 \\
		  &         & GO CC & Mitotic Spindle & $\mathbf{2.4\!\times\!10^{-4}}$ & KIF18A, KIFC1, CDK1, KIF11 \\
		  &         & GO MF & L-glutamine Transmembrane Transporter & $\mathbf{2.7\!\times\!10^{-2}}$ & SLC1A5, SLC38A5 \\
		  &         & KEGG  & p53 signaling pathway & $\mathbf{2.2\!\times\!10^{-4}}$ & RRM2, CASP3, CDK1, PMAIP1 \\
		\cmidrule{2-6}
		  & \multirow{4}{*}{2 (213)} & GO BP & Reg.\ of Cell Population Proliferation & $8.1\!\times\!10^{-2}$ & INSR, IGF1R, PPARG, FOXO4 \\
		  &         & GO CC & Apical Junction Complex & $\mathbf{1.1\!\times\!10^{-3}}$ & CLDN10, RAPGEF2, CGN, CRB3 \\
		  &         & GO MF & Insulin-Like Growth Factor Binding & $\mathbf{2.2\!\times\!10^{-2}}$ & INSR, ITGA6, IGF1R \\
		  &         & KEGG  & Longevity regulating pathway & $\mathbf{1.3\!\times\!10^{-3}}$ & INSR, PPARG, PPARGC1A, IGF1R \\
		\specialrule{0.6pt}{0pt}{0pt}
		\multirow{12}{*}{3} & \multirow{4}{*}{1 (28)}  & GO BP & Actin Filament Organization & $\mathbf{7.2\!\times\!10^{-4}}$ & CNN2, TPM2, DPYSL3, RAC1 \\
		  &         & GO CC & Collagen-Containing ECM & $\mathbf{8.2\!\times\!10^{-4}}$ & SFRP2, COL5A3, COL6A2, BGN \\
		  &         & GO MF & Filamin Binding & $\mathbf{4.8\!\times\!10^{-2}}$ & FBLIM1, DPYSL3 \\
		  &         & KEGG  & Tight junction & $2.0\!\times\!10^{-1}$ & RAC1, MYL12A, JAM3 \\
		\cmidrule{2-6}
		  & \multirow{4}{*}{2 (206)} & GO BP & Response to cGMP & $1.7\!\times\!10^{-1}$ & PDE2A, RAPGEF2 \\
		  &         & GO CC & Apical Junction Complex & $\mathbf{1.5\!\times\!10^{-3}}$ & CLDN10, RAPGEF2, CGN, CRB3 \\
		  &         & GO MF & DNA-binding TF Binding & $\mathbf{3.5\!\times\!10^{-2}}$ & KAT5, BCL2, PPARG, PPARGC1A \\
		  &         & KEGG  & Longevity regulating pathway & $1.7\!\times\!10^{-1}$ & PPARG, PPARGC1A, IGF1R \\
		\cmidrule{2-6}
		  & \multirow{4}{*}{3 (80)}  & GO BP & Reg.\ of CDK Activity & $\mathbf{2.9\!\times\!10^{-4}}$ & CCNA2, CDC6, CDC25C, CDKN3 \\
		  &         & GO CC & Mitotic Spindle & $\mathbf{7.0\!\times\!10^{-7}}$ & KIF18A, KIFC1, CDK1, KIF11 \\
		  &         & GO MF & 5$'$-3$'$ DNA Exonuclease Activity & $\mathbf{1.2\!\times\!10^{-2}}$ & MGME1, EXO1 \\
		  &         & KEGG  & p53 signaling pathway & $\mathbf{2.5\!\times\!10^{-4}}$ & RRM2, CASP3, CDK1, PMAIP1 \\
		\specialrule{0.6pt}{0pt}{0pt}
		\multirow{16}{*}{4} & \multirow{4}{*}{1 (197)} & GO BP & Protein Localization to Cell Periphery & $8.2\!\times\!10^{-2}$ & TMEM88, RAPGEF2, CRB3 \\
		  &         & GO CC & Apical Junction Complex & $\mathbf{8.1\!\times\!10^{-6}}$ & CLDN10, CLDN3, CGN, CRB3 \\
		  &         & GO MF & Acetyltransferase Activity & $1.1\!\times\!10^{-1}$ & NAT8, KAT5, PAFAH2 \\
		  &         & KEGG  & Tight junction & $\mathbf{5.9\!\times\!10^{-3}}$ & CLDN10, CLDN3, ERBB2, CGN \\
		\cmidrule{2-6}
		  & \multirow{4}{*}{2 (16)}  & GO BP & Pos.\ Reg.\ of Cell Adhesion & $9.3\!\times\!10^{-2}$ & SFRP2, LAMB1, RAC1 \\
		  &         & GO CC & Basement Membrane & $7.6\!\times\!10^{-2}$ & CCDC80, PXDN, LAMB1 \\
		  &         & GO MF & Procollagen-Proline Dioxygenase & $5.1\!\times\!10^{-2}$ & P4HA3, P3H3 \\
		  &         & KEGG  & Axon guidance & $5.9\!\times\!10^{-1}$ & RAC1, LIMK1, MYL12A \\
		\cmidrule{2-6}
		  & \multirow{4}{*}{3 (25)}  & GO BP & N/A & N/A & N/A \\
		  &         & GO CC & Collagen-Containing ECM & $3.6\!\times\!10^{-1}$ & COMP, COL5A3, PXDN, BGN \\
		  &         & GO MF & Filamin Binding & $3.1\!\times\!10^{-1}$ & FBLIM1, DPYSL3 \\
		  &         & KEGG  & cGMP-PKG signaling pathway & $1.1\!\times\!10^{-1}$ & INSR, PDE2A, PRKG1 \\
		\cmidrule{2-6}
		  & \multirow{4}{*}{4 (76)}  & GO BP & Microtubule Cytoskeleton Org.\ in Mitosis & $\mathbf{3.1\!\times\!10^{-4}}$ & CDK1, BIRC5, KIF11, NDC80 \\
		  &         & GO CC & Mitotic Spindle & $\mathbf{7.0\!\times\!10^{-6}}$ & KIF18A, KIFC1, CDK1, KIF11 \\
		  &         & GO MF & N/A & N/A & N/A \\
		  &         & KEGG  & p53 signaling pathway & $\mathbf{3.8\!\times\!10^{-3}}$ & RRM2, CASP3, CDK1, PMAIP1 \\
		\specialrule{1.0pt}{0pt}{0pt}
	\end{tabular*}
\end{table*}
}

\kircPathwayTable

The BRCA-transferred embeddings stratify KIRC patients with statistically significant survival differences (Supplementary Table~\ref{tab:kirc_survival_full}). On the DSS endpoint, the overall log-rank $p=3.66\times10^{-8}$. The high-risk Cluster~3 has a hazard ratio of 3.99 against the low-risk Cluster~2, with 95\% CI 2.39 to 6.65 and $p=1.1\times10^{-7}$, and a 36.8\% gap in 5-year DSS rate of 47.0\% versus 83.8\%. The PFI endpoint shows a similar pattern, with overall log-rank $p=1.54\times10^{-6}$; Cluster~3 has HR $= 2.74$ at $p=1.7\times10^{-6}$ and a median PFI of only 1111 days, against 3766 days for Cluster~2. The pairwise DSS log-rank $p$ between Cluster~2 and Cluster~3 reaches $4.18\times10^{-9}$ (complete pairwise tests across $K\in\{2,3,4\}$ for both endpoints are in Supplementary Table~\ref{tab:kirc_pairwise_all}). Figure~\ref{fig:kirc_km} summarizes the DSS (top row) and PFI (bottom row) Kaplan-Meier curves under $K=2,3,4$.

\subsubsection{Pathway Enrichment Analysis}

For the selected $K=3$ solution (Table~\ref{tab:kirc_pathways}), the three clusters map onto distinct biological profiles (ordered below from highest to lowest risk); $K=2$ and $K=4$ results are in the same table.

\begin{itemize}
    \item \textbf{High-risk Cluster~3 (proliferation driven)}. Top enriched terms are in cell cycle and mitosis (Regulation of CDK Activity, adj.\ $p=2.9\times10^{-4}$; KEGG p53 signaling, adj.\ $p=2.5\times10^{-4}$; GO CC Mitotic Spindle, adj.\ $p=7.0\times10^{-7}$). Marker genes include CDK1, BIRC5 (survivin), NDC80, and MKI67, each independently reported as an adverse prognostic indicator in ccRCC: CDK1 kinase activity predicts RCC recurrence \citep{Hongo2014}; BIRC5 has a meta-analysis HR of 2.84 \citep{Xie2016}; NDC80 overexpression drives malignant phenotypes \citep{Hu2023ndc80}. The top 30 differentially expressed genes further include CENPF, MELK, and EXO1.

    \item \textbf{Intermediate-risk Cluster~1 (ECM remodeling)}. Enriched in actin filament organization (adj.\ $p=7.2\times10^{-4}$) and collagen-containing extracellular matrix (adj.\ $p=8.2\times10^{-4}$). Marker genes COL5A3, COL6A2, BGN, SFRP2, and CTHRC1 point to stromal remodeling; RAMP2 supports vascular homeostasis \citep{Tanaka2016} and JAM3 endothelial junctions. The cluster profile is dominated by the mesenchymal and vascular microenvironment remodeling.

    \item \textbf{Low-risk Cluster~2 (epithelial and metabolic maintenance)}. Enriched in epithelial polarity and tight junction structures (Apical Junction Complex, adj.\ $p=1.5\times10^{-3}$; Basolateral Plasma Membrane, adj.\ $p=0.015$). Marker genes PPARG and PPARGC1A (PGC-1$\alpha$) regulate lipid metabolism and mitochondrial biogenesis, FBP1 drives gluconeogenesis, and BCL2 and TIMP3 regulate apoptosis and matrix metalloproteinases. This pattern matches the TCGA ccRCC multi-omics study \citep{TCGA2013kirc}, which identifies metabolic and lipid metabolism features as low risk signatures, with mechanistic support from \citet{Du2017}, who show that HIF drives lipid deposition in ccRCC.
\end{itemize}

\rev{The proliferation axis and epithelial features remain stable across $K=2,3,4$, as the proliferative cluster retains the KEGG p53 and GO CC Mitotic Spindle terms together with CDK1, BIRC5, KIF11, and NDC80, whereas only the secondary epithelial and mesenchymal clusters split or merge (Table~\ref{tab:kirc_pathways}).} \rev{Full marker lists and GO BP plots are provided in Supplementary Table~\ref{tab:kirc_markers_all} and Supplementary Figs.~\ref{fig:kirc_k2_go} to \ref{fig:kirc_k4_go}. Their agreement across resolutions supports a transferable prognostic axis recovered from classification embeddings without using survival outcomes for training.}

\section{Discussion}

CMGL estimates modality reliability separately from classification. Confidence estimation and cross-omics attention therefore address distinct failure modes: one filters noisy modalities, the other combines feature-wise signal across modalities. The ablations agree with this division of labor: confidence estimation helps most when the per-modality signal to noise ratio is low (GBM), while cross-omics attention helps most when no single modality dominates (OV). The two-stage decoupling has a smaller average effect ($-1.6$\%, versus $-3.0$\% for attention), so fusion is the more load-bearing of the two.

Beyond standard benchmark accuracy, the embedding separates a metabolic Class~1 (6.0\%) dominated by tryptophan-kynurenine catabolism and glutathione metabolism (Section~\ref{sec:brca_interpret}). Kynurenine-mediated immune suppression is a known mechanism in the tumor microenvironment, and recovering this signature without any biological supervision suggests that filtering out noisy modalities preserves finer-grained molecular signals in the embedding.

\rev{The model trained on BRCA transfers to KIRC by separating clusters associated with proliferation, mesenchymal programs, and epithelial programs, each with distinct survival profiles (Sections~\ref{sec:kirc_clustering} and~\ref{sec:kirc_survival}).} \rev{Their correspondence to the ccA and ccB framework, together with their contrasting source domain compositions that resemble Luminal A and Basal subtypes, indicates that the proliferation and differentiation axis is preserved across cancers \citep{Brannon2010}.}

\rev{The experiments with modality subsets evaluate view availability across cohorts under consistent inputs, whereas inference with missing modalities for individual patients remains an important extension for clinical deployment.} Graph construction that adapts during training, and joint interpretation across multiple omics layers, are both additional directions for future work.

\FloatBarrier

\section{Conclusion}

We presented CMGL, which uses per-sample uncertainty estimates to guide cross-omics fusion and graph learning. CMGL attains the highest accuracy and Macro-F1 on four MLOmics single-cancer tasks and the pan-cancer task. The BRCA embedding recovers the PAM50 subtypes and isolates a metabolic subtype, and the model trained on BRCA transfers to KIRC without fine tuning, stratifying patients into three clusters matching the ccA and ccB framework. Separating reliability estimation from classification yields accurate and biologically interpretable cross-cancer representations. Missing modality handling and adaptive graph construction are natural extensions.

\section{Conflicts of interest}
The authors declare no competing interests.

\section{Author contributions statement}
Boyang Fan: Methodology, Software, Validation, Writing (Original Draft), Writing (Review \& Editing). Hengchuang Yin: Supervision, Writing (Review \& Editing), Project Administration. Siyu Yi: Formal Analysis, Investigation, Data Curation, Visualization. Yifan Wang: Validation, Investigation, Data Curation, Visualization. Zhicheng Li: Validation, Formal Analysis, Investigation, Visualization. Leijiyu Zhou: Validation, Formal Analysis, Investigation, Data Curation. Jiancheng Lv: Funding Acquisition, Project Administration, Supervision. Wei Ju: Supervision, Writing (Review \& Editing), Project Administration, Funding Acquisition.

\section{Acknowledgments}
The authors thank the anonymous reviewers for their careful reading and constructive comments.

\section{Funding}
This work was supported in part by the National Natural Science Foundation of China under Grants 62306014, 62373348, and 62302495; the Postdoctoral Fellowship Program (Grade A) of CPSF under Grant BX20250376; the Sichuan Science and Technology Program under Grant 2025ZNSFSC1506; the Fundamental Research Funds for the Central Universities under Grant 1082204112K97; the Natural Science Foundation of Xinjiang Uygur Autonomous Region under Grant 2025D01A136; and the Xinjiang Leading Talent Recruitment Program under Grant XJRC-2025-KJ-YJ-CXPT-103.

\section{Data availability}
The datasets used in this study are from the MLOmics benchmark, available at \url{https://github.com/chenzRG/Cancer-Multi-Omics-Benchmark}. The source code for CMGL is available at \url{https://github.com/fanboyang/CMGL} \rev{and \url{https://doi.org/10.5281/zenodo.21921069}. The biological interpretability analyses used the Gene Ontology resource for GO enrichment, available at \url{https://www.geneontology.org/}, the Enrichr gene-set libraries for GO enrichment, available at \url{https://maayanlab.cloud/Enrichr/}, and the KEGG PATHWAY database for pathway enrichment, available at \url{https://www.kegg.jp/en/}.}

\begingroup
\makeatletter
\def\CMGLNewReferenceKeys{Zhang2026prest,Arevalo2017gmu,Baltrusaitis2019multimodal,Dong2023graphnoise,Duan2021multiomics,JU2024106207,Rappoport2019nemo,Wang2014snf,Wang2021defusion}
\@for\CMGLRefKey:=\CMGLNewReferenceKeys\do{\expandafter\def\csname CMGL@newref@\CMGLRefKey\endcsname{1}}
\let\CMGL@original@bibitem\bibitem
\renewcommand{\bibitem}[2][]{%
	\@ifundefined{CMGL@newref@#2}{\color{black}}{\color{revisionred}}%
	\if\relax\detokenize{#1}\relax
		\CMGL@original@bibitem{#2}%
	\else
		\CMGL@original@bibitem[#1]{#2}%
	\fi}
\makeatother
\bibliography{./reference}
\endgroup

\end{document}